%% file: arxiv_camera_ready.tex
\pdfoutput=1
\documentclass[11pt]{article}

\usepackage[preprint]{acl}

\usepackage{times}
\usepackage{latexsym}

\usepackage[T1]{fontenc}
\usepackage[utf8]{inputenc}

\usepackage{microtype}

\usepackage{inconsolata}
\usepackage[most]{tcolorbox}

\usepackage[export]{adjustbox}

\usepackage{graphicx}
\usepackage{booktabs}
\usepackage{multirow}
\usepackage{hyperref}
\usepackage{graphics}
\usepackage{adjustbox}
\usepackage{wrapfig}
\usepackage{caption}
\input{math_commands}

\usepackage[capitalize,noabbrev]{cleveref}  %

\crefname{proposition}{Prop.}{Props.}
\crefname{definition}{Def.}{Defs.}
\crefname{equation}{}{}
\makeatletter
\crefname{section}{Section}{Sections}
\crefname{appendix}{Appendix}{Appendices}
\crefname{subsection}{Section}{Sections}
\crefname{subsubsection}{\S\@gobble}{\S\S}
\makeatother
\crefname{figure}{Figure}{Figures}
\crefname{wrapfigure}{Figure}{Figures}
\crefname{corollary}{Cor.}{Cors.}
\crefname{table}{Table}{Tables}

\usepackage{enumitem}
\setlist[itemize]{noitemsep}

\title{PORT: Preference Optimization on Reasoning Traces}

\author{Salem Lahlou$^*$ \and Abdalgadar Abubaker$^*$ \and Hakim Hacid \\ \texttt{\{salem.lahlou@mbzuai\}@tii.ae}\\
         Technology Innovation Institute, Abu Dhabi}
\author{Salem Lahlou$^{*,\dagger}$ \\
  \small Mohamed bin Zayed University\\
  \small of Artificial Intelligence \\
  \small \texttt{salem.lahlou@mbzuai.ac.ae} \\\And
  Abdalgadar Abubaker$^{*}$ \\
  \small Technology Innovation Institute \\
  \small \texttt{abdalgader.abubaker@tii.ae} \\\And 
  Hakim Hacid \\
  \small Technology Innovation Institute \\
  \small\texttt{hakim.hacid@tii.ae}\\}

\begin{document}

\maketitle
\def\thefootnote{*}\footnotetext{denotes equal contribution}\def\thefootnote{\arabic{footnote}}
\def\thefootnote{$\dagger$}\footnotetext{Work initiated while at Technology Innovation Institute}\def\thefootnote{\arabic{footnote}}
\begin{abstract}
  Preference optimization methods have been successfully applied to improve not only the alignment of large language models (LLMs) with human values, but also specific natural language tasks such as summarization and stylistic continuations. This paper proposes using preference optimization methods on Chain-of-Thought steps in order to improve the mathematical reasoning performances of language models. While the \textit{chosen} answers are obtained from datasets that include reasoning traces, we propose two complementary schemes for generating \textit{rejected} answers: weak LLM prompting, and digit corruption. Our approach leads to increased accuracy on the GSM8K and AQuA-RAT mathematical reasoning benchmarks for Falcon2-11B and Mistral-7B. Additionally, the improved abilities transfer to non-mathematical tasks, including the ARC benchmark and symbolic reasoning challenges. For example, our method can lead to up to relative $8.47\%$ and $18.73\%$ increases in accuracy on the GSM8K and AQuA benchmarks respectively, without any extra annotations. This work suggests that the path towards better language reasoning abilities goes through spending resources on creating high-quality datasets of reasoning traces.
  % \footnote{Our code is publicly available at \url{https://github.com/abdalgader-a/port}.} 
\end{abstract}

\section{Introduction}
In recent years, Large Language Models (LLMs) have been pivotal in democratizing Artificial Intelligence (AI), given their ease of use and impressive abilities in a broad spectrum of tasks. While they have significantly contributed to the striking progress of AI, their success has heavily relied on scaling-up to ever-larger models and datasets. Nonetheless, scaling has not proved sufficient for achieving satisfying results on tasks involving \textit{reasoning}. Reasoning has been a central theme in the history of AI, defining goal posts that push the limits of \textit{intelligence}. The term ``reasoning'' is often used to refer to \textit{informal reasoning}, that ``relies on intuition, experience, and common sense to draw conclusions and solve problems''\citep{huang2022towards}. The limits of scale in eliciting reasoning abilities has been confirmed by analyses in \citet{rae2021scaling,bommasani2021opportunities,cobbe2021training}, amongst others. One reason multi-step reasoning still poses a challenge to LLMs is that the next-word prediction objective used to train them does not explicitly encourage step-by-step reasoning. Chain-of-thought prompting \citep[CoT;][]{wei2022chain}, an augmented prompting strategy, has been shown to improve LLM performances on reasoning tasks, by guiding them to generate sequences of intermediate steps. It should be unsurprising however that solely prompting a language model to ``think step by step'', whether alongside a handful of correct rationales \citep{wei2022chain} or not \citep{kojima2022large}, does not necessarily elicit actual system-2-like \citep{stanovich2000individual,kahneman2003maps} \textit{reasoning} abilities, but at best only mimics humans' thought processes.  Despite claims of the type ``LLMs are decent zero-shot reasoners'' \citep{kojima2022large}, the \textit{emergent} ability of reasoning appears consistently for very large models ($>100B$ parameters) only \citep{wei2022emergent}.
% However, it remains unclear to what extent these performance gains can be attributed to human-like task decomposition or simply the greater computation that additional tokens allow. We show that transformers can use meaningless filler tokens (e.g., '......') in place of a chain of thought to solve two hard algorithmic tasks they could not solve when responding without intermediate tokens'' \citep{pfau2024lets}.
% \textbf{Using verifiers:} \cite{cobbe2021training}  propose training verifiers to evaluate the correctness of model generated solutions on the GSM8K dataset. At test time, they sample a fixed number of candidate solutions and select the solution ranked highest by the verifier. Similarly \cite{shen2021generate} propose to train a language model to jointly generate and rank, thus making the model learns from its own mistakes and making it able to distinguish between correct and incorrect expressions, and show its improved performances on the Math23K \citep{wang-etal-2017-deep} and MAWPS \citep{koncel-kedziorski-etal-2016-mawps} datasets.
\begin{figure*}[h]
    \centering
    \includegraphics[width=1.35\textwidth, center]{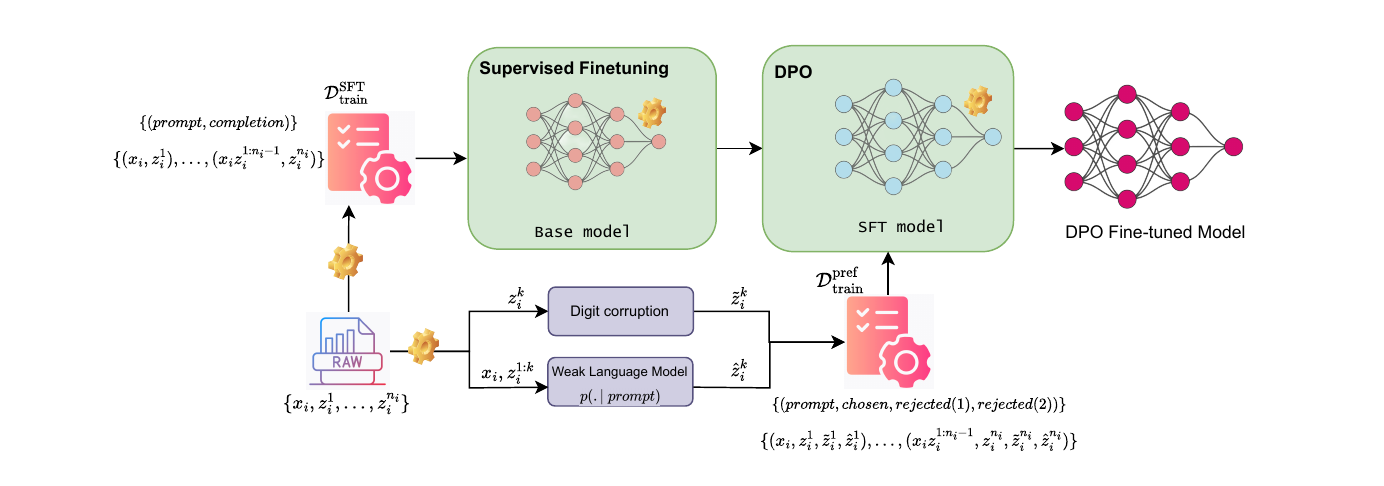}
    \caption{Illustration of the creation process of a preference dataset with two complementary approaches to generate rejected answers. The preference dataset is used to fine-tune a reference model using a Direct Preference Optimization (DPO) or one of its variants, after a supervised fine-tuning (SFT) step. %For a clearer view of the details, please zoom in on the image. %This procedure aims at increasing the likelihood of generating valid reasoning steps and at the same time reducing the likelihood of generating invalid ones.
    }
    \label{fig:main_figure}
\end{figure*}

A major limitation of CoT prompting is its reliance on large models \citep{wei2022chain,kojima2022large}. \citet{ho2022large} propose to bypass this limitation by generating rationales from very large teacher models and using them to fine-tune smaller student models. 
%Similarly, \cite{fu2023specializing} suggests that large models (with more than 100B parameters) have strong modeling power but are spread over a large spectrum of tasks, whereas small models (with less than 10B parameters), can concentrate their capacity on a target task. They thus distill CoT rationales of the GSM8K \citep{cobbe2021training} data from a large teacher model into a smaller model. 
In the same line of work, \citet{uesato2022solving} perform a comprehensive comparison between outcome-based supeversied fine-tuning (SFT), which supervises the final result, and process-based SFT, which supervises the reasoning process, and find that process-based supervision significantly helps language models in mathematical reasoning tasks.
However, solely relying on high-quality rationales is costly as it requires humans or very large language models to generate the reasoning paths. Furthermore, as evidenced by \citet{ni2023learning}, SFT alone tends to make the language model overfit on the rationales seen during training, thus assigning low probabilities to alternative but correct reasoning paths, and as shown in \citet{hong2024orpo}, SFT can still lead the language model to assign high-probabilities to undesired sequences.
% \begin{wrapfigure}{l}{0.75\textwidth}
%   \begin{center}
%     \includegraphics[width=0.75\textwidth]{main_figure.png}
%   \end{center}
%     \caption{Illustration of the process of creation of preference dataset with two complementary approaches to generate rejected answers. The preference dataset is used to fine-tune a reference model using a preference optimization method (e.g. DPO), after a supervised fine-tuning step. This procedure aims at increasing the likelihood of generating valid reasoning steps and at the same time reducing the likelihood of generating invalid ones.}
%     \label{fig:main_figure}
% \end{wrapfigure}

A notable advancement in the development of LLMs is a refinement step that elicits more favorable behaviors. This refinement step is usually performed to align AI systems with human values \citep{gabriel2020artificial,ji2023ai,klingefjord2024human}. The simplest refinement strategy requires a set of demonstrations, or human-made prompt-response examples, and fine-tunes a model on the dataset using supervised fine-tuning. Preference-based approaches on the other hand, rely on datasets of comparisons of potential model outputs. They include reinforcement learning from human feedback 
\citep[RLHF;][]{christiano2017deep,ouyang2022training,bai2022training}, where a reward model is learned and then optimized using the language model as a policy with reinforcement learning algorithms such as the proximal policy optimization algorithm \citep[PPO;][]{schulman2017proximal}. More recently, methods that bypass both the need of explicitly modeling the reward function and the need for online interactions as in RLHF, have become increasingly popular. These include Direct Preference Optimization \citep[DPO;][]{rafailov2024direct}, Identity Preference Optimization \citep[IPO;][]{azar2023ipo}, Sequence Likelihood Calibration with Human Feedback \citep[SLIC;][]{zhao2023slic}, and the prospect theory-based Kahneman-Tversky Optimization \citep[KTO;][]{ethayarajh2024kto}. 
% However, these methods still necessitate that the model first follows an SFT step. More recently however, \cite{hong2024orpo} proposed the Odds Ratio Preference Optimization (ORPO) algorithm that bypasses the need for a separate SFT phase, and combines both SFT and preference optimization into one step, using an adequate loss function. 
Preference optimization techniques have been utilized to improve specific tasks such as summarization and stylistic continuations \citep{ziegler2019finetuning,stiennon2020learning}, but to the best of our knowledge, have never been used to tackle reasoning tasks.
%  \cite{fu2022complexitybased} show that in few-shot CoT prompting, prompts with higher reasoning complexity (i.e. with more reasoning steps) achieve substantially better performance on multi-step reasoning tasks, and use a similar complexity-based criterion to filter-out rationales during decoding. Similarly, \cite{wang2023selfconsistency} propose self-consistency, where multiple rationales are generated during inference and then marginalized out, effectively selecting the most consistent answer, and show that this approach boosts the performance of CoT prompting.

In this paper, we propose to apply preference optimization techniques to chain-of-thought mathematical reasoning. More specifically, we propose two complementary schemes of constructing preference pairs from datasets that include valid mathematical reasoning paths, such as the GSM8K \citep{cobbe2021training} and AQuA \citep{ling-etal-2017-program}. The contributions of the paper are the following:
\begin{itemize}[leftmargin=*,noitemsep,topsep=0pt]
    \item Using Falcon2-11B \citep{malartic2024falcon211b} as our base model, we show that the scheme that relies on corrupting digits to create wrong reasoning steps, can lead to up to $8.47\%$ relative increase in performances on the GSM8K benchmark, and $18.73\%$ on the AQuA benchmark.
    \item We validate \textit{the robustness} of our approach by obtaining favorable results using Mistral-7B \citep{jiang2023mistral} as a base model.
    \item We provide empirical evidence for the transfer abilities of our approach: fine-tuning on mathematical reasoning pairs improves commonsense and symbolic reasoning abilities as well: weak LLM prompting is useful for the ARC benchmark \citep{clark2018think}, and digit corruption is useful for the LastLetterConcat task. \citep{wei2022chain}
    \item We compare the two schemes and various mixtures thereof and provide recommendations of which data mixtures are more susceptible to improve the reasoning abilities.
    \item We compare various preference optimization schemes, and find that DPO leads to better results than its KTO and ORPO variants.
\end{itemize}
Our approach, exemplified by two schemes \textbf{which requires no external data} as illustrated in \cref{fig:main_figure}, suggests that constructing high-quality chain-of-thought datasets that span a wide range of domains holds the promise of improving the emergent reasoning abilities of language models.

% \paragraph{Contributions:} In light of this, we propose in this paper to apply preference optimization techniques to chain-of-thought reasoning. More specifically, we propose two complementary schemes of constructing preference pairs: digit corruption and weak LLM generation. We use DPO on each step of the rationales rather than the full rationales. Given that only very large models exhibit impressive reasoning abilities, we focus on small models and use Falcon2-11B \citep{almazrouei2023falcon} as our base model to fine-tune. To test the robustness of our approach, we evaluate the fine-tuned model using data from the GSM8K dataset \citep{cobbe2021training} only, on the AQuA-RAT \citep{ling-etal-2017-program} and ARC-Challenges \citep{clark2018think} reasoning benchmarks. Our approach is illustrated in \cref{fig:main_figure}.

%\textbf{On GSM8K}: \cite{zhang2024careful} show that many language models have been \textit{contaminated} with the test sets of the standard GSM8K benchmark, and contribute the GSM1K benchmark for more thorough evaluation. We include that in our experimental setup?

\section{PORT: Preference optimization on reasoning traces}
\subsection{Problem setup}
Starting from a finite set of tokens $\gV$, called hereafter the vocabulary, an autoregressive language model can be seen as a collection of probability distributions $\plm$ over $\gV$ conditioned on elements of $
\gV^{\leq \tau} := \bigcup_{t=1}^\tau \gV^t$, i.e. sequences of up to $\tau$ tokens. We assume the existence of an end-of-sentence (EOS) token in $\gV$, denoted $\eos$, that can represent a full stop or a line-break for example. 
\\
To generate text, a pre-trained language model \textit{prompted} with an input $q \in \gV^{\leq \tau}$ is queried autoregressively and samples tokens $s_i \in \gV$, where $s_i \sim \plm(. \mid qs_1\dots s_{i-1})$.
% \begin{equation*}
%     s_1 \sim \plm(.\mid q), \ s_2 \sim \plm(.\mid qs_1), \ s_3 \sim \plm(.\mid qs_1s_2) \dots,
% \end{equation*}
The generation process stops at the first index $k$ for which $s_k = \eos$\footnote{or until the prompt $qs_1\dots s_{k}$ exceeds $\tau$ tokens, but we disregard this case by assuming a very large context window.}. Here, $qs_1\dots s_{i-1}$ refers to the concatenation of the tokens $q, s_1, \dots, s_{i-1}$.
\\
When interacting with language models, and more specifically in CoT reasoning, we are interested in generating sentences $z$, i.e. sequences from $\gV^{\leq T}$ that end with the $\eos$ token, rather than an arbitrary amount of tokens. When prompted with a sentence $x$ or a sequence of sentences $xz^1z^2\dots z^{k-1}$, the language model can therefore autoregressively generate a new sentence $z^k \sim \plm(. \mid xz^1z^2\dots z^{k-1})$.
\\
Given a question $x$ (e.g., a math problem), we define a chain-of-thought as a sequence of $n$ sentences $z^1, \dots, z^n$, where $z^n$ is the final answer. Assuming the existence of a binary function $(x, z) \mapsto \eta(x, z)$ that assesses the correctness of the sentence $z$ to the question $x$, our goal is to tune a pre-trained model $\plm$ to generate a chain $z^1, \dots, z^n$ from a question $x$ such that $\eta(x, z^n)=1$.

\subsection{Proposed approach}
\label{sec:approach}
Our approach first requires access to a dataset of reasoning traces, called a CoT dataset, $\gD_{\mathrm{train}} = \{(x_i, z_i^1, \dots, z_i^{n_i})\}_{i=1}^N$, where each training example includes a \textbf{question} $x_i$, and a \textbf{reasoning trace} (or \textbf{rationale}) comprised of $n_i$ sentences, $z_i^1,\dots,z_i^{n_i}$, of which the last element, $z_i^{n_i}$ is a valid \textbf{answer} to the question $x_i$, i.e., $\eta(x_i, z_i^{n_i}) = 1$. Naturally, the number of steps $n_i$ needed to reach the answer to $x_i$ depends on the question itself.
\\
Such datasets $\gD_{\mathrm{train}}$ are generally human-made. Examples of publicly available reasoning datasets include the arithmetic datasets GSM8K \citep{cobbe2021training}, AQuA-RAT \citep{ling-etal-2017-program}, MAWPS \citep{koncel-kedziorski-etal-2016-mawps} as well as the commonsense reasoning datasets StrategyQA \citep{geva2021aristotle}, Creak \citep{onoe2021creak}, e-SNLI \citep{camburu2018esnli}, ECQA \citep{Aggarwal2021ExplanationsFC}, QASC \citep{khot2019qasc}, QED \citep{lamm2021qed}, Sen-Making \citep{wang-etal-2019-make}. 

\paragraph{SFT data:} From such a dataset, we can construct a dataset $\gD_{\mathrm{train}}^{\mathrm{SFT}}$ of prompt-response pairs, where each example $(x_i, z_i^1, \dots, z_i^{n_i})$ contributes $n_i$ pairs: $(x_i, z_i^1), (x_iz_i^1, z_i^2), \dots, (x_iz_i^1\dots z_i^{n_i-1}, z_i^{n_i})$.
% :
% \begin{align*}
%     \gD_{\mathrm{train}}^{\mathrm{SFT}} = \bigcup_{i=1}^N \bigcup_{k=1}^{n_i} (x_iz_i^{1:k-1}, z_i^k).
% \end{align*}
Such a dataset can be used for supervised fine-tuning, during which the parameters $\vtheta$ of the base language model $\plm$ are updated to minimize the SFT loss:
\begin{align}
    \resizebox{.87\hsize}{!}{$\gL_{\mathrm{SFT}}(\vtheta)= -\frac{1}{N_\mathrm{SFT}} \sum_{i=1}^N \sum_{k=1}^{n_i} p_\vtheta(z_i^k \mid x_iz_i^{1:k-1})$},
    \label{eq:sft_loss}
\end{align}
where $N_\mathrm{SFT} = \sum_{i=1}^N n_i$ = |$\gD_{\mathrm{train}}^{\mathrm{SFT}}$|. In principle, if the data is representative of the target task and if the model generalizes well, the SFT phase should increase the likelihood of \textit{valid reasoning steps}. Put differently, because each $z_i^k$ in the training dataset is a step towards a valid answer $z_i^{n_i}$, then after supervised-fine-tuning, on similar examples, the model should encourage the sentences that unroll the reasoning and help discover a valid answer to the initial question.

\paragraph{Preference data:} From $\gD_{\mathrm{train}}^{\mathrm{SFT}}$, we also construct a preference dataset $\gD_{\mathrm{train}}^{\mathrm{pref}}$, comprised of triplets of the form \texttt{(prompt, chosen, rejected)}. The \textit{prompt} and the \textit{chosen} answers are obtained directly from $\gD_{\mathrm{train}}^{\mathrm{SFT}}$, but for each \textit{prompt} (which is actually either a question or a concatenation of a question and a certain number of initial reasoning steps), we need an invalid reasoning step. Naturally, an arbitrary sequence of tokens would be invalid, but it will provide no useful signal to the model if it is fine-tuned with RLHF or with preference optimization methods such as DPO using such a preference a dataset. Ideally, the \textit{rejected} answers should be almost correct reasoning steps, or contain errors that either a language model or a human are expected to make. Naturally, the \textit{rejected} answers can be obtained using human annotators explicitly asked to generate wrong but close-enough answers. In this work however, we investigate two simple and complementary ways of defining such a dataset:
\begin{itemize}[leftmargin=*,noitemsep,topsep=0pt]
\item \textbf{LLM generation:} For each pair $(x_iz_i^{1:k-1}, z_i^k)$ from $\gD_{\mathrm{train}}^{\mathrm{SFT}}$, we prompt a smaller language model (hereafter also referred to as \textit{weak} LLM) with $x_iz_i^{1:k-1}$ and use the response to define the corresponding \textit{rejected} answer $\hat{z}^k$. By incorporating the resulting triplet in the preference dataset, we naturally incentivize the base model to avoid errors of the type made by the weak LLM. % To avoid cases where the weak LLM produces actual valid step-wise responses,
    %we propose to either filter out the generated responses $\hat{z}^k$ that are judged similar to $z^k$ by a classifier obtained with zero-shot in-context learning from another LLM, or 
    %we propose to further corrupt $\hat{z}^k$ by changing the digits randomly, as explained above. 
This process can be used to generate multiple rejected answers $\hat{z}^k$ per prompt $x_iz_i^{1:k-1}$.
\item \textbf{Digit corruption:} In datasets that involve mathematical reasoning, most reasoning steps $z^k$ include digits. Without modifying any non-digit character of $z^k$, we replace each digit with one from $0$ to $9$ with equal probability. Similarly, this approach can be used to generate multiple rejected answers $\tilde{z}^k$ per prompt $x_iz_i^{1:k-1}$.
\end{itemize}
An illustration of this dataset creation process is provided in \cref{tab:dpo-examples} of \cref{sec:dataset_examples}.
\\
After an SFT phase where $\plm$ is fine-tuned into $p_{\mathrm{SFT}}$ using $\gD_{\mathrm{train}}^{\mathrm{SFT}}$ by minimizing the loss in \cref{eq:sft_loss}, any preference optimization method can be used on  $\gD_{\mathrm{train}}^{\mathrm{pref}}$. For instance, DPO \citep{rafailov2024direct} fine-tunes $p_{\mathrm{SFT}}$ into a model $p_\vtheta$ that minimizes the following loss:
\begin{align}
    \label{eq:dpo_loss}
    \gL_{\mathrm{DPO}}(\vtheta) = - \E_{(x, y_w, y_l)\sim \gD_{\mathrm{train}}^{\mathrm{pref}}} \left[l(x,y_w,y_l;\vtheta)\right],
\end{align}
where
\begin{align}
     l(x,y_w,y_l;\vtheta) = \log &\sigma \left( \beta \log \frac{p_\vtheta(y_w \mid x)}{p_{\mathrm{SFT}}(y_w \mid x)} - \right.\nonumber\\  &\left.\beta \log \frac{p_\vtheta(y_l \mid x)}{p_{\mathrm{SFT}}(y_l \mid x)} \right).\label{eq:indiv_dpo_loss}
\end{align}
Here $\sigma$ is the sigmoid function and $\beta$ is a scaling hyperparameter.
% As an alternative to using two separate SFT and preference optimization phases, ORPO \citep{hong2024orpo}, a more recent approach, combines both steps and replaces the loss in \cref{eq:indiv_dpo_loss} with one that combines

% \begin{align}
% \label{eq:indiv_orpo_loss}
%      l(x,&y_w,y_l;\vtheta) = - \lambda \log \sigma \left( \log \frac{p_\vtheta(y_w \mid x)}{1 - p_\vtheta(y_w \mid x)} - \right. \nonumber \\ &\left. \log \frac{p_\vtheta(y_l \mid x)}{1 - p_\vtheta(y_l \mid x)} \right) - \log p(y_w \mid x),   
% \end{align}
% where $\lambda$ is a weighing hyperparameter.
%To illustrate the process of building preference triplets, we provide in TODO a full example obtained form a specific instance of the GSM8K CoT dataset \citep{cobbe2021training} in Appendix TODO.

\section{Experiments}
\label{sec:experiments}
The goal of the experiments presented in this section are threefold. First, we \textbf{empirically investigate the proposed approach}, and show that unlike SFT, \textbf{it is less prone to task overfitting}. Then, we \textbf{compare the two schemes} above for constructing rejected answers, along with \textbf{different combinations thereof}. Next, we \textbf{investigate the effect of the preference optimization method} by comparing DPO to some of its variants. Finally, we \textbf{validate the robustness} of our method to both the base model and the training dataset. 

\paragraph{Evaluation:} To assess the approach, we need to evaluate the models on informal reasoning tasks, for which chains of thoughts can help reach the valid answers. Our main evaluation task is the GSM8K test dataset \citep{cobbe2021training}, that contains $1319$ high quality grade school math word problems. To assess the transfer abilities of our approach, we also consider the three following evaluation datasets:
\begin{itemize}[leftmargin=*,noitemsep,topsep=0pt]
    \item The Algebra Question Answering with Rationales dataset%\footnote{https://github.com/google-deepmind/AQuA}
    \citep[AQuA;][]{ling-etal-2017-program}, which is a harder math word problem that includes approximately $100,000$ algebraic word problems, each presented with a rationale leading to one of five multiple-choice options (A to E). We use the accompanying test set of $254$ examples for evaluation.
    \item The AI2's Reasoning Challenge%\footnote{https://allenai.org/data/arc}
    \citep[ARC;][]{clark2018think} which is a commonsense reasoning benchmark covering multiple science subjects. The questions are split into \textit{Easy} and \textit{Challenge} sets. Questions in the Challenge set cannot be solved with retrieval or co-occurence methods. Each question admits one valid answer amongst a set of typically four options. We solely focus on the \textbf{Challenge} part.%There is no ground-truth chain-of-thought reasoning provided for this dataset.
    We use the test set of the ARC-Challenge set, that consists of $1172$ examples, for evaluation.
    \item The LastLetterConcat dataset \citep{wei2022chain} which is a symbolic reasoning task where the goal is to join together the last letters of individual words. The dataset contains a total of $500$ examples.
\end{itemize}
More specifically, we use the Language Model Evaluation Harness \citep{eval-harness} to calculate the accuracy (between $0$ and $1$) of the tested models. To elicit the desired CoT behavior, we add few-shot examples from the train set that contain rationales to each question to be evaluated, extract from the generated text the proposed answer, and compare it to the ground truth. We report our results below as percentages. We use 5-shot examples for GSM8K, AQuA, and LastLetterConcat. For ARC, we use 25-shot examples, but given that no rationale is provided in the train set, we use GPT-4 \citep{openai2023gpt4} to construct plausible rationales, and filter them out manually. The used prompt is provided in \cref{sec:used_prompts}.

\paragraph{Base model:} As a base model, we use the newly released pre-trained Falcon2-11B
% \footnote{\url{https://huggingface.co/tiiuae/falcon-11B}} 
\citep{malartic2024falcon211b}
for all our experiments, except in \cref{sec:mistral}, where we confirm that our method is agnostic to the base model by using Mistral-7B \citep{jiang2023mistral}. 
 
\paragraph{Training data:} As previously mentioned, our method requires using a CoT dataset. For all our experiments, except in \cref{sec:using_aqua}, we use the GSM8K \textbf{train} dataset \citep{cobbe2021training}, that consists of $7473$ examples, given that it contains solution steps, in order to construct the SFT and preference datasets $\gD_{\mathrm{train}}^{\mathrm{SFT}}$ and $\gD_{\mathrm{train}}^{\mathrm{pref}}$, as explained in \cref{sec:approach}. Example tuples $(x_i, z_i^1, \dots, z_i^{n_i})$ from this dataset are provided in \cref{sec:dataset_examples}. It is important to note that throughout the training and evaluation process, we only use the training set, without any extra data or human annotation.

\subsection{Supervised fine-tuning}
\label{sec:exp_sft}
From the GSM8K training dataset, we construct the SFT dataset $\gD_{\mathrm{train}}^{\mathrm{SFT}}$ as described in \cref{sec:approach}. The train set of GSM8K consists of $7473$ examples, with an average of $4.57$ reasoning step per example, leading to to an SFT dataset of $34197$ examples. We then fine-tune the based model on this dataset using low-rank adaptation \citep[LoRA;][]{hu2021lora} for efficient parameter updates, processing each example 3 times. The learning rate used in $1.4 \times 10^{-5}$, and the batch size is 16. For LoRA, we use rank 64 matrices and a scaling parameter $\alpha=16$. It is noteworthy that GSM8K examples contain calculation annotations (between <<>>, as shown in the examples provided in \cref{sec:dataset_examples}). These annotations can be used to call external tools (e.g., python scripts or calculators) to perform calculations, rather than asking the LLM to perform the calculation. While we made no such usage of external tools, we tried both keeping and removing the annotations from the text before SFT, and found no significant difference in terms of performance. We thus decided to process the dataset without annotations.
\\
Details about the choice of the hyperparameters are provided in \cref{sec:training_details}.

\paragraph{Results:} In addition to testing the fine-tuned model on the GSM8K's test set, we assess SFT's out-of-distribution generalization, on the harder math word problem AQuA, and on the non-mathematical tasks ARC and LastLetterConcat. We report the accuracies in \cref{tab:base_vs_sft_vs_dpo}. As expected, and as confirmed by other studies \citep{uesato2022solving}, fine-tuning the model on the reasoning steps helps improve the performances on questions requiring reasoning that come from the same distribution. The performances on AQuA and ARC-Challenge drop after the SFT stage, \textbf{confirming the overfitting issues of SFT}, and their limited generalization to unseen examples \citep{ni2023learning}. This is also confirmed by an additional experiment shown in \cref{tab:aqua_n_epochs} in \cref{sec:additional_results}, where we reduce the number of training epochs (on GSM8K) and observe better performances on AQuA.

In the next subsections, we investigate whether preference optimization algorithms can lead to even further performance boosts on the three evaluation tasks.
\begin{table}[t]
        \centering
        \resizebox{\linewidth}{!}{
            \begin{tabular}{lllll}
                \toprule
                \textbf{Model} & GSM8K & AQuA & ARC & LastLetterConcat \\
                \midrule
                Base model &  $54.66$ & $31.50$ & $76.11$ & $16.67$  \\
                \midrule
                SFT & $55.43$ & $30.71$ & $75.60$ & $17.34$   \\
                DPO (ours) & $\mathbf{58.91}$ & $\mathbf{35.04}$ & $76.02$ & $\mathbf{18.67}$\tiny{\textcolor{teal}{($+12\%$)}}\\
                \bottomrule
            \end{tabular}
            }
        \caption{Accuracy (in percentage) of the base, SFT, and DPO models on the three considered tasks. For both SFT and DPO, the Falcon2-11B base model is fine-tuned on datasets obtained from the GSM8K training set. The rejected answers for DPO are obtained using digit corruption, as explained in \cref{sec:dpo_digitcorr} \label{tab:base_vs_sft_vs_dpo}}
\vspace*{-3mm}
\end{table}

\subsection{Preference optimization with digit corruption}
\label{sec:dpo_digitcorr}
From $\gD_{\mathrm{train}}^{\mathrm{SFT}}$, we construct a preference dataset $\gD_{\mathrm{train}}^{\mathrm{pref}}$ using digit corruption as explained in \cref{sec:approach}. Given the stochasticity of the digit corruption approach, we ensure that the rejected answers are indeed invalid, by repeatedly generating reasoning steps until they differ from the ground truth reasoning steps. For reasoning steps that do not include digits, we simply do not include them in the preference dataset.

We fine-tune the SFT model on the obtained preference dataset using DPO with a scale factor $\beta=0.2$, with the same LoRA configuration as SFT. We use the AdamW optimizer \citep{losh2019decoupled} with a learning rate of $8 \times 10^{-6}$ along with a linear schedule for the learning rate. This choice of hyperparameters is explained in \cref{sec:training_details}.

\paragraph{Results:} We report the accuracies post DPO tuning in \cref{tab:base_vs_sft_vs_dpo}. The significant performance increase in GSM8K (a relative $7.77\%$) shows how merely corrupting digits to create \textit{rejected} reasoning steps improves the mathematical reasoning abilities of Falcon2-11B. Our approach helps boost performances on the AQuA task, with a relative increase of $14.41\%$, and on the LastLetterConcat task, with a relative increase of $12\%$, even without using any example from the AQuA train set or from LastLetterConcat during training. These results clearly indicate that, unlike SFT, DPO fine-tuning using digit corruption to construct rejected answers \textbf{instills reasoning skills in the base model}.  We note however, that there is no benefit on the ARC-Challenge task. We suspect that it is because it does not require the same type of skills as GSM8K and AQuA. In \cref{sec:dpo_weakllm}, we investigate whether other schemes could boost ARC performances.

\subsection{Preference optimization using weak LLMs}
\label{sec:dpo_weakllm}
Unlike \cref{sec:dpo_digitcorr}, when constructing $\gD_{\mathrm{train}}^{\mathrm{pref}}$ using weak LLM generation, as described in \cref{sec:approach}, there are a few parameters to take into account: which weak LLM to use? how to prompt said LLM? how to post-process the resulting sequences? 
\\
We first consider the instruct version of the Gemma model \citep{team2024gemma}, Gemma-2B-it, to generate answers. We use the prompt provided in \cref{sec:used_prompts}. We then filter out the responses that do not start with ``Next step: '', and simply do not create the corresponding triplet in the preference dataset. The generation stops at the first line-break or full stop. We also consider the larger Llama-7B \citep{touvron2023llama} and its chat version, to assess the effect of the weak LLM size. 
\\
When using a weak LLM to generate rejected answers, it is not unlikely that the LLM outputs valid reasoning steps, in which case, including the resulting triplet in the preference dataset might hurt generalization of the resulting model. We experimented with the robust version of DPO \citep{chowdhury2024provably}, which accounts for the ambiguity in the preferences, but that did not result in improved performances. We therefore consider to corrupt the digits of the generated sequences similar to the digit corruption scheme alone. In \cref{fig:llama_variants} of \cref{sec:additional_results}, we study the effect of post-generation digit corruption, and find that digit corruption is essential for downstream tasks. We also compare using the chat version of Llama-7B with the prompt template of \cref{sec:used_prompts} to using its base version with few-shot examples only, and find that using the base version yields to better performances. 
\\
Lastly, we consider an \textbf{iterative approach}, where we use the Falcon2-11B fine-tuned with DPO as described in \cref{sec:dpo_digitcorr} as a weak LLM. We report in \cref{tab:comparison_of_weak_llms} the accuracies on the three tasks, using the three weak LLMs with post-generation digit corruption. While DPO with the weak LLM scheme leads to improved results on the math word problems, it does not perform as well as the simpler digit corruption scheme, but it is noteworthy that with Llama-7B and the iterative approach, the performance on ARC-Challenge improves over the base model. This suggests that \textbf{larger models} used as weak LLMs are more likely to generate rejected answers that are informative enough for DPO to \textbf{lead to better models}.

% \begin{table}[h]
%         \centering
%             \begin{tabular}{llllcccc}
%                 \toprule
%                 \textbf{Weak LLM} & GSM8K & AQuA & ARC \\
%                 \midrule
%                 Base model &  $54.66$ & $31.50$ & $76.11$  \\
%                 \midrule
%                 Gemma-2B-it  & $53.68$ & $29.92$ & $75.94$  \\
%                 Llama-7B & $\mathbf{56.10}$ & $30.71$ & $\mathbf{77.05}$  \\
%                 iterative & $55.65$ & $\mathbf{33.46}$ & $76.28$  \\
%                 \bottomrule
%             \end{tabular}
%         \caption{Accuracy (in percentage) of the DPO-finetuned Falcon-11B using three different models as weak LLMs. \textbf{iterative} corresponds to Falcon2-11B fine-tuned with DPO as per \cref{sec:dpo_digitcorr}. To construct the preference dataset, we generate sequences with the LLMs, then we corrupt the digits as described in \cref{sec:approach}. \label{tab:comparison_of_weak_llms}}
% \end{table}

\begin{table}[h]
        \centering
        \resizebox{\linewidth}{!}{
            \begin{tabular}{llll}
                \toprule
                \textbf{Model} & GSM8K & AQuA & ARC \\
                \midrule
                Base model &  $54.66$ & $31.50$ & $76.11$  \\
                \midrule
                \textbf{DPO - effect of weak LLM choice}\\
                Gemma-2B-it  & $53.68$ & $29.92$ & $75.94$  \\
                Llama-7B & $\mathbf{56.10}$ & $30.71$ & $\mathbf{77.05}$  \\
                iterative & $55.65$ & $\mathbf{33.46}$ & $76.28$  \\
                \midrule
                \textbf{DPO - effect of preference data size}\\
                (digit corruption) x 3  & $\mathbf{59.29}$\tiny{\textcolor{teal}{($+8.47\%$)}} & $33.07$ & $76.79$ \\
                (Gemma-2B-it) x 3 & $51.40$ & $\mathbf{35.04}$ & $76.45$ \\
                (Llama-7B) x 3 & $54.51$ & $29.58$ & $76.19$ \\
                Llama-7B + digit corruption & $56.55$ & $32.68$ & $77.47$ \\
                (Llama-7B + digit corruption) x 3 & $56.48$ & $30.31$ & $\mathbf{77.70}$\tiny{\textcolor{teal}{($+2\%$)}} \\
                \bottomrule
            \end{tabular}
            }
        \caption{Accuracy (in percentage) of the DPO-finetuned Falcon-11B using different schemes for rejected answer generation. ``(\texttt{scheme}) x 3'' means that the preference dataset contains 3 rejected answers per chosen answer, obtained by \texttt{scheme}. ``\texttt{scheme1} + \texttt{scheme2}'' means that it contains 2 rejected answers per chosen answer, obtained by concatenating two datasets obtained from \texttt{scheme1} and \texttt{scheme2} respectively. \texttt{iterative} corresponds to Falcon2-11B fine-tuned with DPO as per \cref{sec:dpo_digitcorr}. \label{tab:comparison_of_weak_llms}}
\end{table}

\subsection{Increasing the size of the preference dataset}
A natural question at this point is to consider the effect of the size of the preference dataset on the resulting model fine-tuned with DPO. Given that our proposed approach allows us to generate arbitrarily many wrong reasoning steps per valid reasoning step (e.g., we can corrupt the digits in many ways, and prompt weak LLMs multiple times), we can construct preference datasets with triplets \texttt{(prompt, chosen, rejected)} that contain redundant \texttt{(prompt, chosen)} pairs, with different \texttt{rejected} answers. We thus consider using three rejected answers for the digit corruption, Gemma-2B-it, and Llama-7B experiments. We also consider fine-tuning on a dataset consisting of both digit corrupted answers and Llama-7B-generated answers (themselves digit corrupted), and a dataset containing three times as many rejected answers. The source dataset is GSM8K.
\\
We report in \cref{tab:comparison_of_weak_llms} the results of the different schemes on GSM8K, AQuA, ARC. 
%Strikingly, simply tripling the number of rejected answers for the digit corruption scheme leads to a an accuracy of $59.29\%$ on the GSM8K task, which represents a relative increase of $8.47\%$ over the base performances. Additionally, mixing different sources of rejected answers (Llama-7B and digit corruption), can lead to increased performances on ARC (up to $\mathbf{77.70\%}$ accuracy), indicating that diversifying the sources might help with generalization to other tasks.
These results suggest that \textbf{increasing the preference dataset size} or mixing has the potential of further improving the reasoning abilities of the base language model, and that \textbf{diversifying the sources of rejected answers} might help with generalization to other tasks. For example, simply tripling the number of rejected answers for the digit corruption scheme leads to a an accuracy of $59.29\%$ on the GSM8K task, which represents a relative increase of $8.47\%$ over the base performances. 

\subsection{Benchmarking DPO and its variants}
While DPO has emerged as the go-to method for preference optimization, several variants \citep{azar2023ipo,ethayarajh2024kto} claim to address some of its shortcomings: overfitting, inefficient learning, memory utilization. In this section, we make use of our constructed preference dataset in \cref{sec:dpo_digitcorr} to further compare DPO to its variants. Unlike \citet{saeidi2024insights}, we also consider ORPO \citep{hong2024orpo} which combines both SFT and preference optimization. We report the accuracies on the GSM8K test dataset in \cref{tab:dpo_vs_variants}. We find that the variants of DPO do not lead to improved performances, even with extensive hyperparameter tuning for each method separately (\cref{sec:training_details}). This confirms the recent observations from the benchmark study in \cite{saeidi2024insights} that \textbf{DPO still outperforms its variants on a variety of tasks}.

% \begin{figure}
%     \centering
%     \includegraphics[width=\linewidth]{img/dpo-variants.pdf}
%     \caption{Comparison of alternatives to DPO. The y-axis represents the accuracy on the GSM8K test dataset.}
%     \label{fig:dpo-variants}
%     \vspace*{-3mm}
% \end{figure}
\begin{table}[t]
        \centering
        % \resizebox{\linewidth}{!}{
            \begin{tabular}{lllll}
                \toprule
                \textbf{Method} & DPO & IPO & KTO & ORPO  \\
                \midrule
                \textbf{Accuracy} & $\mathbf{58.91}$ &  $56.40$ & $54.59$ & $55.42$ \\
                \bottomrule
            \end{tabular}
            % }
        \caption{Comparison of alternatives to DPO - accuracy on the GSM8K test set. The preference dataset is constructed using digit corruption only, as in \cref{sec:dpo_digitcorr}. \label{tab:dpo_vs_variants}}
\vspace*{-3mm}
\end{table}

% \begin{figure}[h]
%     \centering
%     \includegraphics[width=\linewidth]{img/large-datasets.pdf}
%     \caption{Effect of increasing the preference dataset size. The answers generated by the weak LLMs undergo digit corruption as well. The horizontal lines correspond to the base model performances.}
%     \label{fig:larger-dataset}
% \end{figure}

\subsection{Robustness analysis: using Mistral as a base model}
\label{sec:mistral}
\begin{figure}[h]
    \centering
    \includegraphics[width=\linewidth]{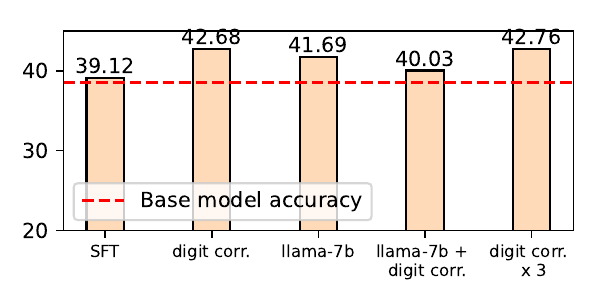}
    \caption{\textbf{Robustness analysis, using Mistral-7B as base model: }GSM8K accuracy - Comparison of different corruption schemes.}
    \label{fig:mistral-results}
    \vspace*{-7mm}
\end{figure}

In this section, we perform some of the experiments above using Mistral-7B \citep{jiang2023mistral} as a base model, rather than Falcon2-11B. We report the results in \cref{fig:mistral-results}, and find that all approaches lead to better performances on the GSM8K benchmark than the base model, which scores $38.51\%$. This experiment confirms the \textbf{robustness of our approach to the base model}, as well as the the \textbf{strength of the digit corruption scheme}.

% \begin{table}[h]
%         \centering
%             \begin{tabular}{llllcccc}
%                 \toprule
%                 \textbf{Weak LLM} & GSM8K & AQuA & ARC \\
%                 \midrule
%                 Gemma-2B-it  & $53.68$ & $29.92$ & $75.94$  \\
%                 Llama-7B & $56.10$ & $30.71$ & $77.05$  \\
%                 iterative & $55.65$ & $33.46$ & $76.28$  \\
%                 \bottomrule
%             \end{tabular}
%         \caption{Accuracy (in percentage) of the base, SFT, and DPO models on the three considered tasks. \textbf{iterative} corresponds to Falcon-11B fine-tuned with DPO as per \cref{sec:dpo_digitcorr}. To construct the preference dataset, we generate sequences with the LLMs, then we corrupt the digits as described in \cref{sec:approach}. \label{tab:comparison_of_weak_llms}}
% \end{table}

\subsection{Robustness analysis: using AQuA as source dataset}
\label{sec:using_aqua}
In this section, we consider using the AQuA training set to create $\gD_{\mathrm{train}}^{\mathrm{SFT}}$ and $\gD_{\mathrm{train}}^{\mathrm{pref}}$. We use Falcon2-11B as a base model. For SFT and DPO, we fine-tune with the same hyper-pararameters as the ones we found best for the experiments on GSM8K. We report the results in \cref{tab:aqua-results}. The results confirm that using the AQuA training set is more helpful for the AQuA benchmark ($18.73\%$ relative increase) than using the GSM8K training set. This experiment also confirms the \textbf{robustness of our approach to the training set}, as well as the the \textbf{strength of the digit corruption scheme}.

\begin{table}[!h]
%\begin{wraptable}{l}{11cm}
        \centering
        \resizebox{\linewidth}{!}{
        % \begin{minipage}{\textwidth}
            \begin{tabular}{llllcccc}
                \toprule
                \textbf{Model} & GSM8K & AQuA & ARC \\
                \midrule
                \small{Base model} &  $54.66$ & $31.50$ & $76.11$  \\
                \midrule
                \small{SFT on AQUA} & $54.89$ & $31.50$ & $75.68$\\
                \small{DPO - digit corr.} & $\mathbf{57.70}$ & $\mathbf{37.40}$\tiny{\textcolor{teal}{($+18.73\%$)}} & $\mathbf{76.88}$  \\
                % \small{DPO - digit corr. x 3} & $54.51$ & $34.25$ & $77.05$  \\
                \small{DPO - Llama-7B} & $55.57$ & $33.86$ & $76.71$  \\
                % DPO--{Llama7B + digit} corruption & $57.85$ & $31.50$ & $77.05$  \\
                % DPO--{Llama7B + digit} corruption x3 & $54.51$ & $34.25$ & $76.62$  \\
                \midrule                
            \end{tabular}
            }
        % \end{minipage}}
        \caption{\textbf{Robustness analysis, using AQuA for training: }Accuracy of the base, SFT, and DPO models (with different schemes).\label{tab:aqua-results}}
        \vspace*{-5mm}
%\end{wraptable}
\end{table}

\section{Related work}
\paragraph{What is reasoning?} Reasoning can be thought of as the process of logically and systematically analyzing information, drawing on evidence and past experiences to form conclusions or make decisions \citep{mchugh2018reasoning}. Using the taxonomy of \citet{huang2022towards}, reasoning can be either \textit{deductive} (a conclusion is drawn based on the truth of the premises), \textit{inductive} (a conclusion is drawn based on observations or evidence), or \textit{abductive} (a conclusion is drawn based on the best explanation for a given set of observations). \citet{bronkhorst2020logical} also make the distinction between \textit{formal} reasoning, akin to what is used in mathematics, in which a fixed set of rules is followed, and \textit{informal} reasoning that is less structured, and is akin to what is used in everyday life.

\paragraph{Reasoning in LLMs:} Mathematics, science, and code  benchmarks \citep{austin2021program,hendrycks2021measuring,liang2023holistic,clark2018think} are becoming increasingly popular to study the emergent reasoning abilities of language models trained on next token prediction. Chain-of-Thought prompting \citep{wei2022chain} and related techniques such as Tree-of-Thought \citep{yao2024tree} and Graph-of-Thought \citep{besta2024graph} have shown to improve language model performances on reasoning tasks, simply by prompting them to generate intermediate computations required for solving the problems. It is not clear however whether the improved performances brought about by chain-of-thought prompting are due specifically to human-like task decomposition, or more generally to the increased computation that additional tokens allow \citep{pfau2024lets}. An orthogonal direction for boosting language model performances on reasoning tasks is reasoning-enhanced training. For example, \citet{lewkowycz2022solving,taylor2022galactica,chen2021evaluating} show that training or fine-tuning LLMs on datasets containing scientific, math, or code data helps improve downstream performances on reasoning tasks. Another line of work \citep{zelikman2022star,huang2022large,gulcehre2023reinforced,yuan2023scaling,singh2023beyond,hosseini2024v} consists of using LLMs to self-improve their reasoning abilities via bootstrapping, where rationales generated by the model that lead to the correct answer are further used to fine-tune the model itself. Aligned with this direction, and more closely related to our work, \citet{ni2023learning} propose to use intermediate steps as supervision signal. \citet{lightman2023lets} conduct a systematic comparison between process supervision (feedback on intermediate steps) and outcome supervision (feedback on final results) for training models on mathematical reasoning, finding that process supervision leads to significantly better performance on the MATH dataset. Another popular set of approaches make use of verifiers, that classify or score reasoning traces \citep{cobbe2021training,uesato2022solving}. As an example of such approaches, GRACE \citep{khalifa2023grace} trains a discriminator with a contrastive loss over correct and incorrect steps, and uses it when generating answers to questions requiring reasoning, to score next-step candidates based on their correctness.

\paragraph{Preference optimization:} To make the most out of a preference dataset, Reinforcement learning with human feedback commonly applies the Bradley-Terry model
\citep{bradley1952rank} to train a reward model that scores instances, and use it to fine-tune the language model to maximize the score of the reward model for the prefered responses using algorithms such as PPO \citep{schulman2017proximal}. More recently, advances in offline methods such as DPO \citep{rafailov2024direct} and its variants \citep{azar2023ipo,zhao2023slic,cai2023ulma,ethayarajh2024kto} that directly align the language models without the need for an explicit reward function, have proven successful in practice. These methods however require an SFT phase to achieve convergence to desired results \citep{rafailov2024direct,tunstall2023zephyr}. ORPO \citep{hong2024orpo} on the other hand, bypasses the need for the multi-stage process, and uses a loss that combines both supervised fine-tuning and preference optimization. In our work, we use these methods as part of the pipeline and compare them thoroughly. Concurrent works \citep{pang2024iterative,lai2024stepdpo} have also applied preference optimization techniques on reasoning data. Our work differs from these concurrent works in both methodology and scope. While \citet{pang2024iterative} focuses on iteratively optimizing between competing CoT candidates and \citet{lai2024stepdpo} proposes step-level preference optimization requiring fine-grained process supervision, our work introduces two novel and complementary schemes for generating rejected answers (weak LLM prompting and digit corruption) that require no additional annotations or external data. Furthermore, we provide a comprehensive empirical study comparing different preference optimization variants (DPO, IPO, KTO, ORPO) for reasoning tasks. Unlike these works which focus solely on mathematical reasoning, we demonstrate that our approach transfers to non-mathematical tasks, including commonsense and symbolic reasoning, suggesting broader implications for improving general reasoning abilities in language models.

\section{Conclusion}

We considered the question of using preference optimization to boost reasoning abilities of language models. More specifically, we proposed two different schemes for constructing preference datasets of reasoning steps from datasets that include valid reasonign traces. We showed that by using DPO on these datasets, we are able to improve the reasoning abilities of Falcon2-11B and Mistral-7B, even on tasks unseen during training. We also compared DPO to several of its variants. Our work suggests that constructing high-quality reasoning traces datasets can boost general informal reasoning abilities.
% built from a classifier obtained with zero-shot in-context learning from another LLM, that filters out generated responses that are similar to the ground truth.
\section*{Limitations}
We considered two schemes for wrong reasoning step generation in this work: digit corruption, and LLM generation. There are several other ways that could be considered. For instance, it could be beneficial to consider prompting an LLM to slightly tweak the ground-truth reasoning steps until they become wrong.  We leave the study of other schemes to future work, along with scaling to models over 11B. Additionally, when using the weak LLM scheme, there is an overhead incurred when creating the dataset. Finally, our work has focused on mathematical reasoning, and future work should explore using other sources of reasoning data. Perhaps mixing between different sources of data, as suggested by recent work from \citet{chung2024scaling}, could lead to improved abilities in out-of-distribution reasoning benchmarks.

\section*{Acknowledgments}
The authors would like to thank Mohamed El Amine Seddik and Reda Alami for fruitful discussions that helped improve this manuscript.

\bibliography{sample}

\newpage
\clearpage
\appendix
\section{Dataset examples}
\label{sec:dataset_examples}
We provide in the following box two examples from the training dataset of GSM8K, the main source of data used in our experiments. We also provide in \cref{tab:dpo-examples} the preference triplets obtained from \textbf{one} example from the GSM8K (a third one). The ground truth rationale for this particular example contains three sentences, and thus contributes three examples to the preference dataset.
\begin{tcolorbox}[colback=yellow!5!white,colframe=yellow!50!black,
  colbacktitle=yellow!75!black,title=Examples from the GSM8K training dataset]
  \textbf{Question:} John puts \$25 in his piggy bank every month for 2 years to save up for a vacation. He had to spend \$400 from his piggy bank savings last week to repair his car. How many dollars are left in his piggy bank?\\
  \textbf{Rationale:} He saved money for 2 years, which is equal to 12 x 2 = <<12*2=24>>24 months. The amount of money he saved is \$25*24 = \$<<25*24=600>>600. But he spent some money so there is \$600 - \$400 = <<600-400=200>>200 left. \#\#\#\# 200.
  \tcblower
  \textbf{Question:} Five coaster vans are used to transport students for their field trip. Each van carries 28 students, 60 of which are boys. How many are girls?\\
  \textbf{Rationale:} There are a total of 5 vans x 28 students = <<5*28=140>>140 students. If 60 are boys, then 140 - 60 = <<140-60=80>>80 of these students are girls. \#\#\#\# 80
\end{tcolorbox}

\section{Used prompts}
\label{sec:used_prompts}
In the following box, we provide the prompt used to generate plausible rationales for the 25 few shot examples of ARC-Challenge, using GPT-4.
\begin{tcolorbox}[colback=yellow!5!white,colframe=yellow!50!black,
  colbacktitle=yellow!75!black,title=GPT-4 prompt for ARC rationale generation]
  You are expert grade-school science teacher. Given the following question, provide justification for the answer.\\
  Question: \texttt{question}. Answer Choices: \texttt{options}. Answer: ... The Answer is \texttt{answer letter}.\\
  You need to add a two to three sentences rationale before ``The answer is \texttt{answer letter}'', justifying the correct answer.
\end{tcolorbox}
Next, we provide the template used to prompt the weak LLMs to generate rejected answers to create the preference dataset for \cref{sec:dpo_weakllm}.
\begin{tcolorbox}[colback=yellow!5!white,colframe=yellow!50!black,
  colbacktitle=yellow!75!black,title=Prompt template for weak LLM generation]
  You are an obedient assistant. Your task is to reason about the following question. Write only the next step of the reasoning chain. Your answer should include exactly one following reasoning step and has to be exactly one sentence long! The answer should start with "Next step: ". Here are two examples:\\
\begin{verbatim}
Question: {prompt_example_1}
Next step: {first_step_example_1}
Next step: {second_step_example_1}
Next step: {third_step_example_1}
Next step: {fourth_step_example_1}
Next step: {final_answer_example_1}

Question: {prompt_example_2}
Next step: {first_step_example_2}
Next step: {second_step_example_2}
Next step: {third_step_example_2}
Next step: {fourth_step_example_2}
Next step: {fifth_step_example_2}
Next step: {sixth_step_example_2}
Next step: {final_answer_example_2}

Question: {prompt}
Next step: {first_step_ground_truth}
...
\end{verbatim}
\end{tcolorbox}
\begin{table*}[h!]
    \begin{tabular}{|p{5.7cm}|p{2.8cm}|p{2.8cm}|p{2.8cm}|}
    \cline{1-4}
    \textbf{prompt} & \textbf{chosen} & \textbf{rejected(1)} & \textbf{rejected(2)} \\ \cline{1-4}
     \small{Natalia sold clips to 48 of her friends in April, and then she sold half as many clips in May. How many clips did Natalia sell altogether in April and May?} & \small{Natalia sold 48/2 = 24 clips in May.} & \small{Natalia sold 32/4 = 19 clips in May.} & \small{Natalia sold 48 clips in April, and half as many clips in May, which is 24 clips} \\  \cline{1-4}
     \small{Natalia sold clips to 48 of her friends in April, and then she sold half as many clips in May. How many clips did Natalia sell altogether in April and May? Natalia sold 48/2 = 24 clips in May.} & \small{Natalia sold 48+24 = 72 clips altogether in April and May.} & \small{Natalia sold 25+98 = 12 clips altogether in April and May.}  & \small{Natalia sold 24 clips in April.} \\ \cline{1-4}
     \small{Natalia sold clips to 48 of her friends in April, and then she sold half as many clips in May. How many clips did Natalia sell altogether in April and May? Natalia sold 48/2 = 24 clips in May. Natalia sold 48+24 = 72 clips altogether in April and May.}  & \small{The solution to the problem is 72.}  & \small{The solution to the problem is 13.}  & \small{Natalia sold 24 clips in April, so she sold 24 clips in May.}  \\ \cline{1-4}
    \end{tabular}
    \caption{Example of preference dataset obtained from the \textbf{question} ``Natalia sold clips to 48 of her friends in April, and then she sold half as many clips in May. How many clips did Natalia sell altogether in April and May?'', and the corresponding \textbf{rationale} ``Natalia sold $48/2 = 24$ clips in May. Natalia sold $48+24 = 72$ clips altogether in April and May. The solution to the problem is $72$.'' This example is obtained from the GSM8K dataset \citep{cobbe2021training}. The \textbf{chosen} column represents steps from ground-truth rationale, \textbf{rejected (1)} are examples obtained by digit corruption, and \textbf{rejected (2)} are examples obtained by prompting the Llama-2-7B-chat model \citep{touvron2023llama} 
}
    \label{tab:dpo-examples}
\end{table*}

\section{Training details}
\label{sec:training_details}
\paragraph{Number of epochs used for SFT:} An important hyperparameter when doing supervised fine-tuning on small datasets is the number of times each example is processed. We optimized this hyperparameter independently using the smaller Falcon-7B \citep{almazrouei2023falcon} as a base model, with the GSM8K accuracy as a metric, trying values $\{1,2,3,4,5\}$. We ended up using 3 epochs for all subsequent SFT experiments (using Falcon2-11B as a base model).

\paragraph{SFT learning rate:} Similarly, using the GSM8K accuracy as a metric, and with a random search of a handful of learning rates in the range $[10^{-8}, 10^{-4}]$, we ended up using the learning rate of $1.4 \times 10^{-5}$.

\paragraph{SFT batch size:} It is commonly agreed upon that larger batch sizes are more desirable when fine-tuning language models. We used a batch size of 16 as it was the largest that did not lead to memory issues on the GPUs we used.

\paragraph{LoRA parameters:} Given the computational cost of fine-tuning LLMs, we chose not to tune the hyperparameters of LoRA \citep{hu2021lora}, and resorted to using the popular values of $\alpha=16$ and $rank=64$.
 
\paragraph{Optimizer:} For DPO, we used a linear schedule for the learning rate, and first jointly optimized the maximal learning rate and number of warm-up steps for the linear schedule, using RMSProp \citep{tieleman2012lecture} as an optimizer. Optimizing this couple of hyperparameters was done on Falcon-7B using a preference dataset with digit corrupted rejected answers only, with the GSM8K accuracy as a metric. After settling on $10$ warm-up steps, we tuned the maximal learning rate and the optimizer (choosing between RMSProp and AdamW \citep{losh2019decoupled}) on the same (model, task, metric) triplet. We ended up using the AdamW optimizer with a maximal learning rate of $8 \times 10^{-6}$.

\begin{figure*}[t]
    \centering
    \includegraphics[width=1.\textwidth, center]{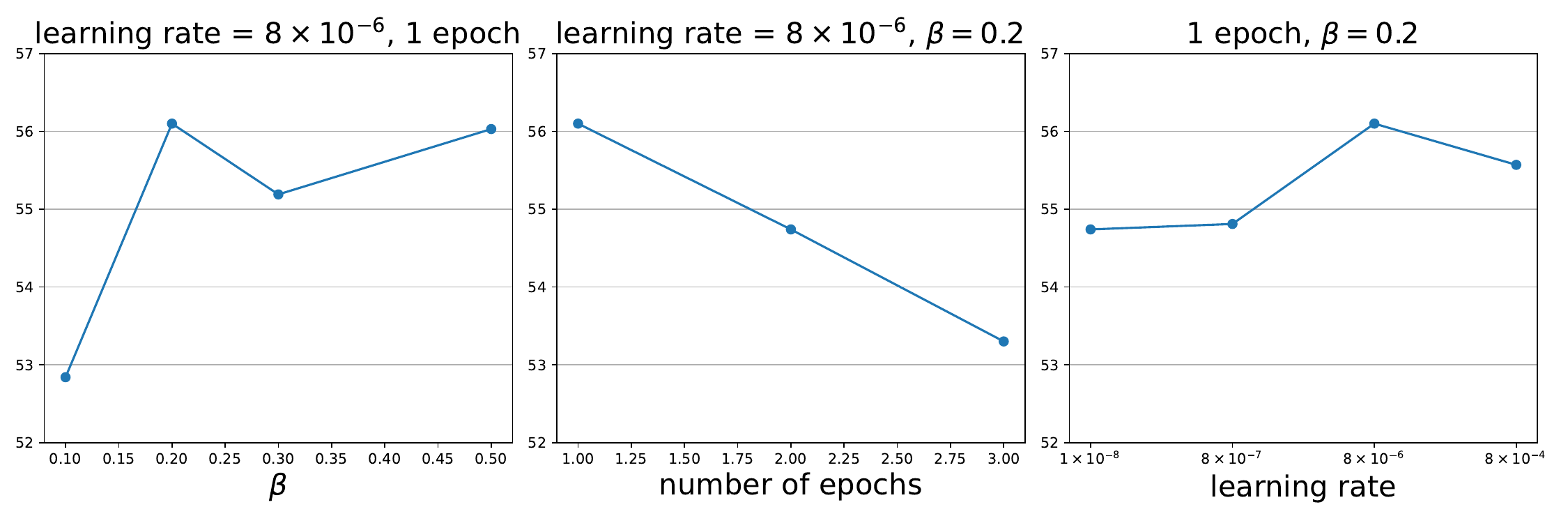}
    \caption{DPO hyperparameter search. The y axis corresponds to the accuracy on the test set of GSM8K.}
    \label{fig:dpo_hp_search}
\end{figure*}

\paragraph{Further DPO hyperparameters:} We further optimized the learning rate, as well as the number of training epochs and the value of the $\beta$ hyperparameter on the preference dataset $\gD_{\mathrm{train}}^{\mathrm{pref}}$ constructed from GSM8K, and using digit-corrupted Llama-7B \citep{touvron2023llama}, as explained in \cref{sec:dpo_weakllm}, to generate wrong answers. Using the resulting model's performance on the GSM8K task, with the evaluation protocol described in \cref{sec:experiments}, we report the results of our hyperparameter sweep in \cref{fig:dpo_hp_search}. This led to a universal choice of $\beta=0.2$, learning rate of $8 \times 10^{-6}$, and number of epochs equal to $1$ for all DPO experiments with Falcon2-11B.

\begin{figure}[h!]
    \centering
    \includegraphics[width=\linewidth]{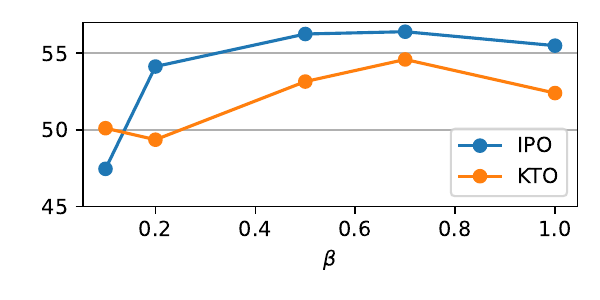}
    \caption{DPO variants hyperparameter search. The y axis corresponds to the accuracy on the test set of GSM8K. The learning rate $8 \times 10^{-6}$ and number of epochs ($1$) used are the same as DPO.}
    \label{fig:dpo_variants_hp_search}
\end{figure}

\paragraph{Hyperparameters for DPO variants:} Similar to DPO, the KTO \citep{ethayarajh2024kto} loss requires the specification of a hyperparameter $\beta$ that controls how far the fine-tuned model drifts from the SFT model. IPO \citep{azar2023ipo} needs a regularization parameter $\tau$, for which the inverse $\tau^{-1}$ is usually denoted by $\beta$ as well. For both methods, the value of $\beta$ is critical and needs to be carefully tuned. We report in \cref{fig:dpo_variants_hp_search} the results of our hyperparameter search. We consider the preference dataset $\gD_{\mathrm{train}}^{\mathrm{pref}}$ constructed from GSM8K, and using digit corruption, as explained in \cref{sec:dpo_digitcorr}, to generate wrong answers.
\\
For ORPO \citep{hong2024orpo}, an important parameter is the weighing hyperparameter $\lambda$ in \cref{eq:indiv_orpo_loss}, that specifies the relative importance of the negative log-likelihood of the chosen answer with respect to the odds ratio part of the loss. We tried the values in the set $\{0.001, 0.005, 0.01, 0.1, 0.2, 0.3\}$ along with learning rates from the set $\{10^{-8}, 8 \times 10^{-8}, 8 \times 10^{-6}\}$, and found that $\beta=0.001$ and $10^{-8}$ as a learning rate lead to the best results, which is what we report in \cref{tab:dpo_vs_variants}.%\cref{fig:dpo-variants}.

\section{Additional Results}
\label{sec:additional_results}
\begin{figure}[h]
    \centering
    \includegraphics[width=\linewidth]{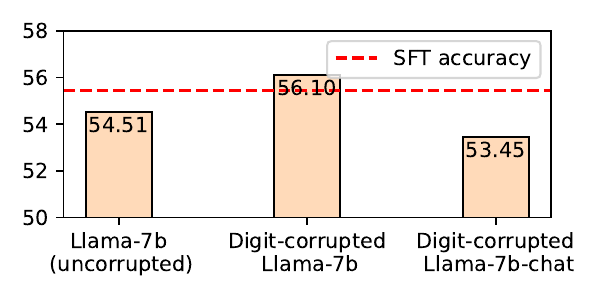}
    \caption{DPO with weak LLM generation for rejected answers. Comparison of different versions of Llama-7B. The y axis corresponds to the accuracy on the test set of GSM8K. The learning rate use is $8 \times 10^{-6}$ and number of epochs is $1$.}
    \label{fig:llama_variants}
\end{figure}

In \cref{tab:aqua_n_epochs}, we study how the number of times each example from the GSM8K training dataset is visited during training affects the downstream performance on the related but different AQuA evaluation task. The table shows that reducing the training time could help dampen the overfitting issues of SFT.
\begin{table}[h]
        \centering
            \begin{tabular}{llllcccc}
                \toprule
                 & 1 epoch & 3 epochs  \\
                \midrule
                AQuA accuracy &  $33.46$ & $30.71$  \\
                \bottomrule
            \end{tabular}
        \caption{Accuracy (in percentage) on the AQuA test dataset of Falcon2-11B fine-tuned on $\gD_{\mathrm{train}}^{\mathrm{SFT}}$ obtained from the GSM8K train dataset, as explained in \cref{sec:exp_sft}. Comparison of the effect of number of epochs. \label{tab:aqua_n_epochs}}
\end{table}
\\
When using a weak LLM to generate rejected answers, it is not unlikely that the LLM outputs valid reasoning steps, in which case, including the resulting triplet in the preference dataset might hurt generalization of the resulting model. We therefore consider to corrupt the digits of the generated sequences similar to the digit corruption scheme alone. In \cref{fig:llama_variants}, we study the effect of post-generation digit corruption, and find that digit corruption is essential for downstream tasks. We also compare using the chat version of Llama-7B with the prompt template of \cref{sec:used_prompts} to using its base version, and find that using the base version yields to better performances.

\subsection{Ablations}
We hypothesized that fine-tuning a language model to predict the next reasoning step only should help improve performances on reasoning benchmarks. Our results in the main paper confirm this hypothesis. However, it is natural to wonder whether using multiple reasoning steps to predict could be beneficial. More specifically, for SFT, we compare our approach (which requires fine-tuning on $(xz^{1:k-1}, z^{k})$ pairs) to fine-tuning on $(xz^{1:k-1}, z^{k:n})$ pairs. Similarly, for DPO, we compare our approach (that requires fine-tuning on $(xz^{1:k-1}, z^{k}, \tilde{z}^k)$ triplets) to fine-tuning on $(xz^{1:k-1}, z^{k:n}, \tilde{z}^kz^{k+1:n})$ triplets. Using Falcon2-11B as a base model, and GSM8K as a data source and for evaluation, we found that with this change, the performance drops from $55.43$ to $54.81$ for SFT, and from $58.30$ to $57.01$ for DPO with digit corruption.
\\
We also tested replacing the inputs $xz^{1:k-1}$ with $uxz^{1:k-1}$, where $u$ is a sequence corresponding to 3-shot examples, and found that while the SFT performance slightly increases to $55.95$, the DPO performance significantly drops to $50.11$.

\end{document}

%% file: math_commands.tex
\usepackage{amsmath,amsfonts,bm}

\def\vtheta{{\bm{\theta}}}

\def\gD{{\mathcal{D}}}

\def\gL{{\mathcal{L}}}

\def\gV{{\mathcal{V}}}

\newcommand{\E}{\mathbb{E}}

\newcommand{\plm}{p_{\mathrm{LM}}}
\newcommand{\eos}{\mathrm{EOS}}

%% file: arxiv_camera_ready.bbl
\begin{thebibliography}{73}
\providecommand{\natexlab}[1]{#1}

\bibitem[{Aggarwal et~al.(2021)Aggarwal, Mandowara, Agrawal, Khandelwal, Singla, and Garg}]{Aggarwal2021ExplanationsFC}
Shourya Aggarwal, Divyanshu Mandowara, Vishwajeet Agrawal, Dinesh Khandelwal, Parag Singla, and Dinesh Garg. 2021.
\newblock \href {https://api.semanticscholar.org/CorpusID:236459873} {Explanations for commonsenseqa: New dataset and models}.
\newblock In \emph{Annual Meeting of the Association for Computational Linguistics}.

\bibitem[{Almazrouei et~al.(2023)Almazrouei, Alobeidli, Alshamsi, Cappelli, Cojocaru, Debbah, Étienne Goffinet, Hesslow, Launay, Malartic, Mazzotta, Noune, Pannier, and Penedo}]{almazrouei2023falcon}
Ebtesam Almazrouei, Hamza Alobeidli, Abdulaziz Alshamsi, Alessandro Cappelli, Ruxandra Cojocaru, Mérouane Debbah, Étienne Goffinet, Daniel Hesslow, Julien Launay, Quentin Malartic, Daniele Mazzotta, Badreddine Noune, Baptiste Pannier, and Guilherme Penedo. 2023.
\newblock The falcon series of open language models.
\newblock \emph{arXiv preprint arXiv: 2311.16867}.

\bibitem[{Austin et~al.(2021)Austin, Odena, Nye, Bosma, Michalewski, Dohan, Jiang, Cai, Terry, Le et~al.}]{austin2021program}
Jacob Austin, Augustus Odena, Maxwell Nye, Maarten Bosma, Henryk Michalewski, David Dohan, Ellen Jiang, Carrie Cai, Michael Terry, Quoc Le, et~al. 2021.
\newblock Program synthesis with large language models.
\newblock \emph{arXiv preprint arXiv:2108.07732}.

\bibitem[{Azar et~al.(2023)Azar, Rowland, Piot, Guo, Calandriello, Valko, and Munos}]{azar2023ipo}
Mohammad~Gheshlaghi Azar, Mark Rowland, Bilal Piot, Daniel Guo, Daniele Calandriello, Michal Valko, and R{\'e}mi Munos. 2023.
\newblock A general theoretical paradigm to understand learning from human preferences.
\newblock \emph{arXiv preprint arXiv:2310.12036}.

\bibitem[{Bai et~al.(2022)Bai, Jones, Ndousse, Askell, Chen, DasSarma, Drain, Fort, Ganguli, Henighan et~al.}]{bai2022training}
Yuntao Bai, Andy Jones, Kamal Ndousse, Amanda Askell, Anna Chen, Nova DasSarma, Dawn Drain, Stanislav Fort, Deep Ganguli, Tom Henighan, et~al. 2022.
\newblock Training a helpful and harmless assistant with reinforcement learning from human feedback.
\newblock \emph{arXiv preprint arXiv:2204.05862}.

\bibitem[{Besta et~al.(2024)Besta, Blach, Kubicek, Gerstenberger, Podstawski, Gianinazzi, Gajda, Lehmann, Niewiadomski, Nyczyk et~al.}]{besta2024graph}
Maciej Besta, Nils Blach, Ales Kubicek, Robert Gerstenberger, Michal Podstawski, Lukas Gianinazzi, Joanna Gajda, Tomasz Lehmann, Hubert Niewiadomski, Piotr Nyczyk, et~al. 2024.
\newblock Graph of thoughts: Solving elaborate problems with large language models.
\newblock In \emph{Proceedings of the AAAI Conference on Artificial Intelligence}, volume~38, pages 17682--17690.

\bibitem[{Bommasani et~al.(2021)Bommasani, Hudson, Adeli, Altman, Arora, von Arx, Bernstein, Bohg, Bosselut, Brunskill, Brynjolfsson et~al.}]{bommasani2021opportunities}
Rishi Bommasani, Drew~A. Hudson, Ehsan Adeli, Russ Altman, Simran Arora, Sydney von Arx, Michael~S. Bernstein, Jeannette Bohg, Antoine Bosselut, Emma Brunskill, Erik Brynjolfsson, et~al. 2021.
\newblock \href {https://arxiv.org/abs/2108.07258v3} {On the opportunities and risks of foundation models}.
\newblock \emph{arXiv preprint arXiv: 2108.07258}.

\bibitem[{Bradley and Terry(1952)}]{bradley1952rank}
Ralph~Allan Bradley and Milton~E Terry. 1952.
\newblock Rank analysis of incomplete block designs: I. the method of paired comparisons.
\newblock \emph{Biometrika}, 39(3/4):324--345.

\bibitem[{Bronkhorst et~al.(2020)Bronkhorst, Roorda, Suhre, and Goedhart}]{bronkhorst2020logical}
Hugo Bronkhorst, Gerrit Roorda, Cor Suhre, and Martin Goedhart. 2020.
\newblock Logical reasoning in formal and everyday reasoning tasks.
\newblock \emph{International Journal of Science and Mathematics Education}, 18:1673--1694.

\bibitem[{Cai et~al.(2023)Cai, Song, Jiang, Teng, Gu, and Zhang}]{cai2023ulma}
Tianchi Cai, Xierui Song, Jiyan Jiang, Fei Teng, Jinjie Gu, and Guannan Zhang. 2023.
\newblock Ulma: Unified language model alignment with demonstration and point-wise human preference.
\newblock \emph{arXiv preprint arXiv:2312.02554}.

\bibitem[{Camburu et~al.(2018)Camburu, Rockt\"{a}schel, Lukasiewicz, and Blunsom}]{camburu2018esnli}
Oana-Maria Camburu, Tim Rockt\"{a}schel, Thomas Lukasiewicz, and Phil Blunsom. 2018.
\newblock \href {https://proceedings.neurips.cc/paper_files/paper/2018/file/4c7a167bb329bd92580a99ce422d6fa6-Paper.pdf} {e-snli: Natural language inference with natural language explanations}.
\newblock In \emph{Advances in Neural Information Processing Systems}, volume~31. Curran Associates, Inc.

\bibitem[{Chen et~al.(2021)Chen, Tworek, Jun, Yuan, de~Oliveira~Pinto, Kaplan, Edwards, Burda, Joseph, Brockman, Ray, Puri, Krueger, Petrov, Khlaaf, Sastry, Mishkin, Chan, Gray, Ryder, Pavlov, Power, Kaiser, Bavarian, Winter, Tillet, Such, Cummings, Plappert, Chantzis, Barnes, Herbert-Voss, Guss, Nichol, Paino, Tezak, Tang, Babuschkin, Balaji, Jain, Saunders, Hesse, Carr, Leike, Achiam, Misra, Morikawa, Radford, Knight, Brundage, Murati, Mayer, Welinder, McGrew, Amodei, McCandlish, Sutskever, and Zaremba}]{chen2021evaluating}
Mark Chen, Jerry Tworek, Heewoo Jun, Qiming Yuan, Henrique~Ponde de~Oliveira~Pinto, Jared Kaplan, Harri Edwards, Yuri Burda, Nicholas Joseph, Greg Brockman, Alex Ray, Raul Puri, Gretchen Krueger, Michael Petrov, Heidy Khlaaf, Girish Sastry, Pamela Mishkin, Brooke Chan, Scott Gray, Nick Ryder, Mikhail Pavlov, Alethea Power, Lukasz Kaiser, Mohammad Bavarian, Clemens Winter, Philippe Tillet, Felipe~Petroski Such, Dave Cummings, Matthias Plappert, Fotios Chantzis, Elizabeth Barnes, Ariel Herbert-Voss, William~Hebgen Guss, Alex Nichol, Alex Paino, Nikolas Tezak, Jie Tang, Igor Babuschkin, Suchir Balaji, Shantanu Jain, William Saunders, Christopher Hesse, Andrew~N. Carr, Jan Leike, Josh Achiam, Vedant Misra, Evan Morikawa, Alec Radford, Matthew Knight, Miles Brundage, Mira Murati, Katie Mayer, Peter Welinder, Bob McGrew, Dario Amodei, Sam McCandlish, Ilya Sutskever, and Wojciech Zaremba. 2021.
\newblock Evaluating large language models trained on code.
\newblock \emph{arXiv preprint arXiv: 2107.03374}.

\bibitem[{Chowdhury et~al.(2024)Chowdhury, Kini, and Natarajan}]{chowdhury2024provably}
Sayak~Ray Chowdhury, Anush Kini, and Nagarajan Natarajan. 2024.
\newblock Provably robust dpo: Aligning language models with noisy feedback.
\newblock \emph{arXiv preprint arXiv: 2403.00409}.

\bibitem[{Christiano et~al.(2017)Christiano, Leike, Brown, Martic, Legg, and Amodei}]{christiano2017deep}
Paul~F Christiano, Jan Leike, Tom Brown, Miljan Martic, Shane Legg, and Dario Amodei. 2017.
\newblock Deep reinforcement learning from human preferences.
\newblock \emph{Advances in neural information processing systems}, 30.

\bibitem[{Chung et~al.(2024)Chung, Hou, Longpre, Zoph, Tay, Fedus, Li, Wang, Dehghani, Brahma et~al.}]{chung2024scaling}
Hyung~Won Chung, Le~Hou, Shayne Longpre, Barret Zoph, Yi~Tay, William Fedus, Yunxuan Li, Xuezhi Wang, Mostafa Dehghani, Siddhartha Brahma, et~al. 2024.
\newblock Scaling instruction-finetuned language models.
\newblock \emph{Journal of Machine Learning Research}, 25(70):1--53.

\bibitem[{Clark et~al.(2018)Clark, Cowhey, Etzioni, Khot, Sabharwal, Schoenick, and Tafjord}]{clark2018think}
Peter Clark, Isaac Cowhey, Oren Etzioni, Tushar Khot, Ashish Sabharwal, Carissa Schoenick, and Oyvind Tafjord. 2018.
\newblock Think you have solved question answering? try arc, the ai2 reasoning challenge.
\newblock \emph{arXiv preprint arXiv: 1803.05457}.

\bibitem[{Cobbe et~al.(2021)Cobbe, Kosaraju, Bavarian, Chen, Jun, Kaiser, Plappert, Tworek, Hilton, Nakano et~al.}]{cobbe2021training}
Karl Cobbe, Vineet Kosaraju, Mohammad Bavarian, Mark Chen, Heewoo Jun, Lukasz Kaiser, Matthias Plappert, Jerry Tworek, Jacob Hilton, Reiichiro Nakano, et~al. 2021.
\newblock Training verifiers to solve math word problems.
\newblock \emph{arXiv preprint arXiv:2110.14168}.

\bibitem[{et~al.(2023)}]{openai2023gpt4}
OpenAI et~al. 2023.
\newblock Gpt-4 technical report.
\newblock \emph{arXiv preprint arXiv: 2303.08774}.

\bibitem[{Ethayarajh et~al.(2024)Ethayarajh, Xu, Muennighoff, Jurafsky, and Kiela}]{ethayarajh2024kto}
Kawin Ethayarajh, Winnie Xu, Niklas Muennighoff, Dan Jurafsky, and Douwe Kiela. 2024.
\newblock \href {https://arxiv.org/abs/2402.01306v1} {Kto: Model alignment as prospect theoretic optimization}.
\newblock \emph{arXiv preprint arXiv: 2402.01306}.

\bibitem[{Gabriel(2020)}]{gabriel2020artificial}
Iason Gabriel. 2020.
\newblock Artificial intelligence, values, and alignment.
\newblock \emph{Minds and machines}, 30(3):411--437.

\bibitem[{Gao et~al.(2023)Gao, Tow, Abbasi, Biderman, Black, DiPofi, Foster, Golding, Hsu, Le~Noac'h, Li, McDonell, Muennighoff, Ociepa, Phang, Reynolds, Schoelkopf, Skowron, Sutawika, Tang, Thite, Wang, Wang, and Zou}]{eval-harness}
Leo Gao, Jonathan Tow, Baber Abbasi, Stella Biderman, Sid Black, Anthony DiPofi, Charles Foster, Laurence Golding, Jeffrey Hsu, Alain Le~Noac'h, Haonan Li, Kyle McDonell, Niklas Muennighoff, Chris Ociepa, Jason Phang, Laria Reynolds, Hailey Schoelkopf, Aviya Skowron, Lintang Sutawika, Eric Tang, Anish Thite, Ben Wang, Kevin Wang, and Andy Zou. 2023.
\newblock \href {https://doi.org/10.5281/zenodo.10256836} {A framework for few-shot language model evaluation}.

\bibitem[{Geva et~al.(2021)Geva, Khashabi, Segal, Khot, Roth, and Berant}]{geva2021aristotle}
Mor Geva, Daniel Khashabi, Elad Segal, Tushar Khot, D.~Roth, and Jonathan Berant. 2021.
\newblock \href {https://doi.org/10.1162/tacl_a_00370} {Did aristotle use a laptop? a question answering benchmark with implicit reasoning strategies}.
\newblock \emph{Transactions of the Association for Computational Linguistics}.

\bibitem[{Gulcehre et~al.(2023)Gulcehre, Paine, Srinivasan, Konyushkova, Weerts, Sharma, Siddhant, Ahern, Wang, Gu et~al.}]{gulcehre2023reinforced}
Caglar Gulcehre, Tom~Le Paine, Srivatsan Srinivasan, Ksenia Konyushkova, Lotte Weerts, Abhishek Sharma, Aditya Siddhant, Alex Ahern, Miaosen Wang, Chenjie Gu, et~al. 2023.
\newblock Reinforced self-training (rest) for language modeling.
\newblock \emph{arXiv preprint arXiv:2308.08998}.

\bibitem[{Hendrycks et~al.(2021)Hendrycks, Burns, Kadavath, Arora, Basart, Tang, Song, and Steinhardt}]{hendrycks2021measuring}
Dan Hendrycks, Collin Burns, Saurav Kadavath, Akul Arora, Steven Basart, Eric Tang, Dawn Song, and Jacob Steinhardt. 2021.
\newblock \href {https://openreview.net/forum?id=7Bywt2mQsCe} {Measuring mathematical problem solving with the {MATH} dataset}.
\newblock In \emph{Thirty-fifth Conference on Neural Information Processing Systems Datasets and Benchmarks Track (Round 2)}.

\bibitem[{Ho et~al.(2022)Ho, Schmid, and Yun}]{ho2022large}
Namgyu Ho, Laura Schmid, and Se-Young Yun. 2022.
\newblock \href {https://doi.org/10.48550/arXiv.2212.10071} {Large language models are reasoning teachers}.
\newblock \emph{Annual Meeting of the Association for Computational Linguistics}.

\bibitem[{Hong et~al.(2024)Hong, Lee, and Thorne}]{hong2024orpo}
Jiwoo Hong, Noah Lee, and James Thorne. 2024.
\newblock \href {https://arxiv.org/abs/2403.07691v2} {Orpo: Monolithic preference optimization without reference model}.
\newblock \emph{arXiv preprint arXiv: 2403.07691}.

\bibitem[{Hosseini et~al.(2024)Hosseini, Yuan, Malkin, Courville, Sordoni, and Agarwal}]{hosseini2024v}
Arian Hosseini, Xingdi Yuan, Nikolay Malkin, Aaron Courville, Alessandro Sordoni, and Rishabh Agarwal. 2024.
\newblock V-star: Training verifiers for self-taught reasoners.
\newblock \emph{arXiv preprint arXiv:2402.06457}.

\bibitem[{Hu et~al.(2021)Hu, Shen, Wallis, Allen-Zhu, Li, Wang, Wang, and Chen}]{hu2021lora}
Edward~J. Hu, Yelong Shen, Phillip Wallis, Zeyuan Allen-Zhu, Yuanzhi Li, Shean Wang, Lu~Wang, and Weizhu Chen. 2021.
\newblock Lora: Low-rank adaptation of large language models.
\newblock \emph{arXiv preprint arXiv: 2106.09685}.

\bibitem[{Huang et~al.(2022)Huang, Gu, Hou, Wu, Wang, Yu, and Han}]{huang2022large}
Jiaxin Huang, Shixiang~Shane Gu, Le~Hou, Yuexin Wu, Xuezhi Wang, Hongkun Yu, and Jiawei Han. 2022.
\newblock Large language models can self-improve.
\newblock \emph{arXiv preprint arXiv:2210.11610}.

\bibitem[{Huang and Chang(2022)}]{huang2022towards}
Jie Huang and Kevin Chen-Chuan Chang. 2022.
\newblock Towards reasoning in large language models: A survey.
\newblock \emph{arXiv preprint arXiv:2212.10403}.

\bibitem[{Ji et~al.(2023)Ji, Qiu, Chen, Zhang, Lou, Wang, Duan, He, Zhou, Zhang et~al.}]{ji2023ai}
Jiaming Ji, Tianyi Qiu, Boyuan Chen, Borong Zhang, Hantao Lou, Kaile Wang, Yawen Duan, Zhonghao He, Jiayi Zhou, Zhaowei Zhang, et~al. 2023.
\newblock Ai alignment: A comprehensive survey.
\newblock \emph{arXiv preprint arXiv:2310.19852}.

\bibitem[{Jiang et~al.(2023)Jiang, Sablayrolles, Mensch, Bamford, Chaplot, de~las Casas, Bressand, Lengyel, Lample, Saulnier, Lavaud, Lachaux, Stock, Scao, Lavril, Wang, Lacroix, and Sayed}]{jiang2023mistral}
Albert~Q. Jiang, Alexandre Sablayrolles, Arthur Mensch, Chris Bamford, Devendra~Singh Chaplot, Diego de~las Casas, Florian Bressand, Gianna Lengyel, Guillaume Lample, Lucile Saulnier, Lélio~Renard Lavaud, Marie-Anne Lachaux, Pierre Stock, Teven~Le Scao, Thibaut Lavril, Thomas Wang, Timothée Lacroix, and William~El Sayed. 2023.
\newblock Mistral 7b.
\newblock \emph{arXiv preprint arXiv: 2310.06825}.

\bibitem[{Kahneman(2003)}]{kahneman2003maps}
Daniel Kahneman. 2003.
\newblock Maps of bounded rationality: Psychology for behavioral economics.
\newblock \emph{American economic review}, 93(5):1449--1475.

\bibitem[{Khalifa et~al.(2023)Khalifa, Logeswaran, Lee, Lee, and Wang}]{khalifa2023grace}
Muhammad Khalifa, Lajanugen Logeswaran, Moontae Lee, Ho~Hin Lee, and Lu~Wang. 2023.
\newblock \href {https://doi.org/10.18653/v1/2023.findings-emnlp.1022} {Grace: Discriminator-guided chain-of-thought reasoning}.
\newblock \emph{Conference on Empirical Methods in Natural Language Processing}.

\bibitem[{Khot et~al.(2019)Khot, Clark, Guerquin, Jansen, and Sabharwal}]{khot2019qasc}
Tushar Khot, Peter Clark, Michal Guerquin, Peter~Alexander Jansen, and Ashish Sabharwal. 2019.
\newblock \href {https://doi.org/10.1609/AAAI.V34I05.6319} {Qasc: A dataset for question answering via sentence composition}.
\newblock \emph{AAAI Conference on Artificial Intelligence}.

\bibitem[{Klingefjord et~al.(2024)Klingefjord, Lowe, and Edelman}]{klingefjord2024human}
Oliver Klingefjord, Ryan Lowe, and Joe Edelman. 2024.
\newblock \href {https://arxiv.org/abs/2404.10636v2} {What are human values, and how do we align ai to them?}
\newblock \emph{arXiv preprint arXiv: 2404.10636}.

\bibitem[{Kojima et~al.(2022)Kojima, Gu, Reid, Matsuo, and Iwasawa}]{kojima2022large}
Takeshi Kojima, Shixiang~Shane Gu, Machel Reid, Yutaka Matsuo, and Yusuke Iwasawa. 2022.
\newblock Large language models are zero-shot reasoners.
\newblock \emph{Advances in neural information processing systems}, 35:22199--22213.

\bibitem[{Koncel-Kedziorski et~al.(2016)Koncel-Kedziorski, Roy, Amini, Kushman, and Hajishirzi}]{koncel-kedziorski-etal-2016-mawps}
Rik Koncel-Kedziorski, Subhro Roy, Aida Amini, Nate Kushman, and Hannaneh Hajishirzi. 2016.
\newblock \href {https://doi.org/10.18653/v1/N16-1136} {{MAWPS}: A math word problem repository}.
\newblock In \emph{Proceedings of the 2016 Conference of the North {A}merican Chapter of the Association for Computational Linguistics: Human Language Technologies}, pages 1152--1157, San Diego, California. Association for Computational Linguistics.

\bibitem[{Lai et~al.(2024)Lai, Tian, Chen, Yang, Peng, and Jia}]{lai2024stepdpo}
Xin Lai, Zhuotao Tian, Yukang Chen, Senqiao Yang, Xiangru Peng, and Jiaya Jia. 2024.
\newblock Step-dpo: Step-wise preference optimization for long-chain reasoning of llms.
\newblock \emph{arXiv preprint arXiv: 2406.18629}.

\bibitem[{Lamm et~al.(2021)Lamm, Palomaki, Alberti, Andor, Choi, Soares, and Collins}]{lamm2021qed}
Matthew Lamm, Jennimaria Palomaki, Chris Alberti, Daniel Andor, Eunsol Choi, Livio~Baldini Soares, and Michael Collins. 2021.
\newblock Qed: A framework and dataset for explanations in question answering.
\newblock \emph{Transactions of the Association for computational Linguistics}, 9:790--806.

\bibitem[{Lewkowycz et~al.(2022)Lewkowycz, Andreassen, Dohan, Dyer, Michalewski, Ramasesh, Slone, Anil, Schlag, Gutman-Solo et~al.}]{lewkowycz2022solving}
Aitor Lewkowycz, Anders Andreassen, David Dohan, Ethan Dyer, Henryk Michalewski, Vinay Ramasesh, Ambrose Slone, Cem Anil, Imanol Schlag, Theo Gutman-Solo, et~al. 2022.
\newblock Solving quantitative reasoning problems with language models.
\newblock \emph{Advances in Neural Information Processing Systems}, 35:3843--3857.

\bibitem[{Liang et~al.(2023)Liang, Bommasani, Lee, Tsipras, Soylu, Yasunaga, Zhang, Narayanan, Wu, Kumar et~al.}]{liang2023holistic}
Percy Liang, Rishi Bommasani, Tony Lee, Dimitris Tsipras, Dilara Soylu, Michihiro Yasunaga, Yian Zhang, Deepak Narayanan, Yuhuai Wu, Ananya Kumar, et~al. 2023.
\newblock Holistic evaluation of language models.
\newblock \emph{Transactions on Machine Learning Research}.

\bibitem[{Lightman et~al.(2023)Lightman, Kosaraju, Burda, Edwards, Baker, Lee, Leike, Schulman, Sutskever, and Cobbe}]{lightman2023lets}
Hunter Lightman, V.~Kosaraju, Yura Burda, Harrison Edwards, Bowen Baker, Teddy Lee, J.~Leike, John Schulman, I.~Sutskever, and K.~Cobbe. 2023.
\newblock \href {https://doi.org/10.48550/arXiv.2305.20050} {Let's verify step by step}.
\newblock \emph{International Conference on Learning Representations}.

\bibitem[{Ling et~al.(2017)Ling, Yogatama, Dyer, and Blunsom}]{ling-etal-2017-program}
Wang Ling, Dani Yogatama, Chris Dyer, and Phil Blunsom. 2017.
\newblock \href {https://doi.org/10.18653/v1/P17-1015} {Program induction by rationale generation: Learning to solve and explain algebraic word problems}.
\newblock In \emph{Proceedings of the 55th Annual Meeting of the Association for Computational Linguistics (Volume 1: Long Papers)}, pages 158--167, Vancouver, Canada. Association for Computational Linguistics.

\bibitem[{Loshchilov and Hutter(2019)}]{losh2019decoupled}
Ilya Loshchilov and Frank Hutter. 2019.
\newblock \href {https://openreview.net/forum?id=Bkg6RiCqY7} {Decoupled weight decay regularization}.
\newblock In \emph{7th International Conference on Learning Representations, {ICLR} 2019, New Orleans, LA, USA, May 6-9, 2019}. OpenReview.net.

\bibitem[{Malartic et~al.(2024)Malartic, Chowdhury, Cojocaru, Farooq, Campesan, Djilali, Narayan, Singh, Velikanov, Boussaha, Al-Yafeai, Alobeidli, Qadi, Seddik, Fedyanin, Alami, and Hacid}]{malartic2024falcon211b}
Quentin Malartic, Nilabhra~Roy Chowdhury, Ruxandra Cojocaru, Mugariya Farooq, Giulia Campesan, Yasser Abdelaziz~Dahou Djilali, Sanath Narayan, Ankit Singh, Maksim Velikanov, Basma El~Amel Boussaha, Mohammed Al-Yafeai, Hamza Alobeidli, Leen~Al Qadi, Mohamed El~Amine Seddik, Kirill Fedyanin, Reda Alami, and Hakim Hacid. 2024.
\newblock Falcon2-11b technical report.
\newblock \emph{arXiv preprint arXiv: 2407.14885}.

\bibitem[{McHugh and Way(2018)}]{mchugh2018reasoning}
Conor McHugh and Jonathan Way. 2018.
\newblock What is reasoning?
\newblock \emph{Mind}, 127(505):167--196.

\bibitem[{Ni et~al.(2023)Ni, Inala, Wang, Polozov, Meek, Radev, and Gao}]{ni2023learning}
Ansong Ni, Jeevana~Priya Inala, Chenglong Wang, Alex Polozov, Christopher Meek, Dragomir Radev, and Jianfeng Gao. 2023.
\newblock \href {https://openreview.net/forum?id=4D4TSJE6-K} {Learning math reasoning from self-sampled correct and partially-correct solutions}.
\newblock In \emph{The Eleventh International Conference on Learning Representations}.

\bibitem[{Onoe et~al.(2021)Onoe, Zhang, Choi, and Durrett}]{onoe2021creak}
Yasumasa Onoe, Michael~J.Q. Zhang, Eunsol Choi, and Greg Durrett. 2021.
\newblock Creak: A dataset for commonsense reasoning over entity knowledge.
\newblock \emph{NeurIPS Datasets and Benchmarks}.

\bibitem[{Ouyang et~al.(2022)Ouyang, Wu, Jiang, Almeida, Wainwright, Mishkin, Zhang, Agarwal, Slama, Ray, Schulman, Hilton, Kelton, Miller, Simens, Askell, Welinder, Christiano, Leike, and Lowe}]{ouyang2022training}
Long Ouyang, Jeffrey Wu, Xu~Jiang, Diogo Almeida, Carroll Wainwright, Pamela Mishkin, Chong Zhang, Sandhini Agarwal, Katarina Slama, Alex Ray, John Schulman, Jacob Hilton, Fraser Kelton, Luke Miller, Maddie Simens, Amanda Askell, Peter Welinder, Paul~F Christiano, Jan Leike, and Ryan Lowe. 2022.
\newblock \href {https://proceedings.neurips.cc/paper_files/paper/2022/file/b1efde53be364a73914f58805a001731-Paper-Conference.pdf} {Training language models to follow instructions with human feedback}.
\newblock In \emph{Advances in Neural Information Processing Systems}, volume~35, pages 27730--27744. Curran Associates, Inc.

\bibitem[{Pang et~al.(2024)Pang, Yuan, Cho, He, Sukhbaatar, and Weston}]{pang2024iterative}
Richard~Yuanzhe Pang, Weizhe Yuan, Kyunghyun Cho, He~He, Sainbayar Sukhbaatar, and Jason Weston. 2024.
\newblock Iterative reasoning preference optimization.
\newblock \emph{arXiv preprint arXiv: 2404.19733}.

\bibitem[{Pfau et~al.(2024)Pfau, Merrill, and Bowman}]{pfau2024lets}
Jacob Pfau, William Merrill, and Samuel~R. Bowman. 2024.
\newblock \href {https://arxiv.org/abs/2404.15758v1} {Let's think dot by dot: Hidden computation in transformer language models}.
\newblock \emph{arXiv preprint arXiv: 2404.15758}.

\bibitem[{Rae et~al.(2021)Rae, Borgeaud, Cai, Millican, Hoffmann, Song, Aslanides, Henderson, Ring, Young et~al.}]{rae2021scaling}
Jack~W Rae, Sebastian Borgeaud, Trevor Cai, Katie Millican, Jordan Hoffmann, Francis Song, John Aslanides, Sarah Henderson, Roman Ring, Susannah Young, et~al. 2021.
\newblock Scaling language models: Methods, analysis \& insights from training gopher.
\newblock \emph{arXiv preprint arXiv:2112.11446}.

\bibitem[{Rafailov et~al.(2024)Rafailov, Sharma, Mitchell, Manning, Ermon, and Finn}]{rafailov2024direct}
Rafael Rafailov, Archit Sharma, Eric Mitchell, Christopher~D Manning, Stefano Ermon, and Chelsea Finn. 2024.
\newblock Direct preference optimization: Your language model is secretly a reward model.
\newblock \emph{Advances in Neural Information Processing Systems}, 36.

\bibitem[{Saeidi et~al.(2024)Saeidi, Verma, and Baral}]{saeidi2024insights}
Amir Saeidi, Shivanshu Verma, and Chitta Baral. 2024.
\newblock Insights into alignment: Evaluating dpo and its variants across multiple tasks.
\newblock \emph{arXiv preprint arXiv: 2404.14723}.

\bibitem[{Schulman et~al.(2017)Schulman, Wolski, Dhariwal, Radford, and Klimov}]{schulman2017proximal}
John Schulman, Filip Wolski, Prafulla Dhariwal, Alec Radford, and Oleg Klimov. 2017.
\newblock Proximal policy optimization algorithms.
\newblock \emph{arXiv preprint arXiv:1707.06347}.

\bibitem[{Singh et~al.(2023)Singh, Co-Reyes, Agarwal, Anand, Patil, Garcia, Liu, Harrison, Lee, Xu, Parisi, Kumar, Alemi, Rizkowsky, Nova, Adlam, Bohnet, Elsayed, Sedghi, Mordatch, Simpson, Gur, Snoek, Pennington, Hron, Kenealy, Swersky, Mahajan, Culp, Xiao, Bileschi, Constant, Novak, Liu, Warkentin, Qian, Bansal, Dyer, Neyshabur, Sohl-Dickstein, and Fiedel}]{singh2023beyond}
Avi Singh, John~D. Co-Reyes, Rishabh Agarwal, Ankesh Anand, Piyush Patil, Xavier Garcia, Peter~J. Liu, James Harrison, Jaehoon Lee, Kelvin Xu, Aaron Parisi, Abhishek Kumar, Alex Alemi, Alex Rizkowsky, Azade Nova, Ben Adlam, Bernd Bohnet, Gamaleldin Elsayed, Hanie Sedghi, Igor Mordatch, Isabelle Simpson, Izzeddin Gur, Jasper Snoek, Jeffrey Pennington, Jiri Hron, Kathleen Kenealy, Kevin Swersky, Kshiteej Mahajan, Laura Culp, Lechao Xiao, Maxwell~L. Bileschi, Noah Constant, Roman Novak, Rosanne Liu, Tris Warkentin, Yundi Qian, Yamini Bansal, Ethan Dyer, Behnam Neyshabur, Jascha Sohl-Dickstein, and Noah Fiedel. 2023.
\newblock Beyond human data: Scaling self-training for problem-solving with language models.
\newblock \emph{arXiv preprint arXiv: 2312.06585}.

\bibitem[{Stanovich et~al.(2000)Stanovich, West, and Alder}]{stanovich2000individual}
Keith~E Stanovich, Richard~F West, and JE~Alder. 2000.
\newblock Individual differences in reasoning: Implications for the rationality debate?-open peer commentary-three fallacies.
\newblock \emph{Behavioral and Brain Sciences}, 23(5):665--665.

\bibitem[{Stiennon et~al.(2020)Stiennon, Ouyang, Wu, Ziegler, Lowe, Voss, Radford, Amodei, and Christiano}]{stiennon2020learning}
Nisan Stiennon, Long Ouyang, Jeff Wu, Daniel~M. Ziegler, Ryan~J. Lowe, Chelsea Voss, Alec Radford, Dario Amodei, and Paul Christiano. 2020.
\newblock Learning to summarize from human feedback.
\newblock \emph{Neural Information Processing Systems}.

\bibitem[{Taylor et~al.(2022)Taylor, Kardas, Cucurull, Scialom, Hartshorn, Saravia, Poulton, Kerkez, and Stojnic}]{taylor2022galactica}
Ross Taylor, Marcin Kardas, Guillem Cucurull, Thomas Scialom, Anthony Hartshorn, Elvis Saravia, Andrew Poulton, Viktor Kerkez, and Robert Stojnic. 2022.
\newblock Galactica: A large language model for science.
\newblock \emph{arXiv preprint arXiv: 2211.09085}.

\bibitem[{Team et~al.(2024)Team, Mesnard, Hardin, Dadashi, Bhupatiraju, Pathak, Sifre, Rivi{\`e}re, Kale, Love et~al.}]{team2024gemma}
Gemma Team, Thomas Mesnard, Cassidy Hardin, Robert Dadashi, Surya Bhupatiraju, Shreya Pathak, Laurent Sifre, Morgane Rivi{\`e}re, Mihir~Sanjay Kale, Juliette Love, et~al. 2024.
\newblock Gemma: Open models based on gemini research and technology.
\newblock \emph{arXiv preprint arXiv:2403.08295}.

\bibitem[{Tieleman(2012)}]{tieleman2012lecture}
Tijmen Tieleman. 2012.
\newblock Lecture 6.5-rmsprop: Divide the gradient by a running average of its recent magnitude.
\newblock \emph{COURSERA: Neural networks for machine learning}, 4(2):26.

\bibitem[{Touvron et~al.(2023)Touvron, Martin, Stone, Albert, Almahairi, Babaei, Bashlykov, Batra, Bhargava, Bhosale, Bikel, Blecher, Ferrer, Chen, Cucurull, Esiobu, Fernandes, Fu, Fu, Fuller, Gao, Goswami, Goyal, Hartshorn, Hosseini, Hou, Inan, Kardas, Kerkez, Khabsa, Kloumann, Korenev, Koura, Lachaux, Lavril, Lee, Liskovich, Lu, Mao, Martinet, Mihaylov, Mishra, Molybog, Nie, Poulton, Reizenstein, Rungta, Saladi, Schelten, Silva, Smith, Subramanian, Tan, Tang, Taylor, Williams, Kuan, Xu, Yan, Zarov, Zhang, Fan, Kambadur, Narang, Rodriguez, Stojnic, Edunov, and Scialom}]{touvron2023llama}
Hugo Touvron, Louis Martin, Kevin Stone, Peter Albert, Amjad Almahairi, Yasmine Babaei, Nikolay Bashlykov, Soumya Batra, Prajjwal Bhargava, Shruti Bhosale, Dan Bikel, Lukas Blecher, Cristian~Canton Ferrer, Moya Chen, Guillem Cucurull, David Esiobu, Jude Fernandes, Jeremy Fu, Wenyin Fu, Brian Fuller, Cynthia Gao, Vedanuj Goswami, Naman Goyal, Anthony Hartshorn, Saghar Hosseini, Rui Hou, Hakan Inan, Marcin Kardas, Viktor Kerkez, Madian Khabsa, Isabel Kloumann, Artem Korenev, Punit~Singh Koura, Marie-Anne Lachaux, Thibaut Lavril, Jenya Lee, Diana Liskovich, Yinghai Lu, Yuning Mao, Xavier Martinet, Todor Mihaylov, Pushkar Mishra, Igor Molybog, Yixin Nie, Andrew Poulton, Jeremy Reizenstein, Rashi Rungta, Kalyan Saladi, Alan Schelten, Ruan Silva, Eric~Michael Smith, Ranjan Subramanian, Xiaoqing~Ellen Tan, Binh Tang, Ross Taylor, Adina Williams, Jian~Xiang Kuan, Puxin Xu, Zheng Yan, Iliyan Zarov, Yuchen Zhang, Angela Fan, Melanie Kambadur, Sharan Narang, Aurelien Rodriguez, Robert Stojnic, Sergey Edunov, and Thomas
  Scialom. 2023.
\newblock Llama 2: Open foundation and fine-tuned chat models.
\newblock \emph{arXiv preprint arXiv: 2307.09288}.

\bibitem[{Tunstall et~al.(2023)Tunstall, Beeching, Lambert, Rajani, Rasul, Belkada, Huang, von Werra, Fourrier, Habib et~al.}]{tunstall2023zephyr}
Lewis Tunstall, Edward Beeching, Nathan Lambert, Nazneen Rajani, Kashif Rasul, Younes Belkada, Shengyi Huang, Leandro von Werra, Cl{\'e}mentine Fourrier, Nathan Habib, et~al. 2023.
\newblock Zephyr: Direct distillation of lm alignment.
\newblock \emph{arXiv preprint arXiv:2310.16944}.

\bibitem[{Uesato et~al.(2022)Uesato, Kushman, Kumar, Song, Siegel, Wang, Creswell, Irving, and Higgins}]{uesato2022solving}
Jonathan Uesato, Nate Kushman, Ramana Kumar, Francis Song, Noah Siegel, Lisa Wang, Antonia Creswell, Geoffrey Irving, and Irina Higgins. 2022.
\newblock \href {https://arxiv.org/abs/2211.14275v1} {Solving math word problems with process- and outcome-based feedback}.
\newblock \emph{arXiv preprint arXiv: 2211.14275}.

\bibitem[{Wang et~al.(2019)Wang, Liang, Zhang, Li, and Gao}]{wang-etal-2019-make}
Cunxiang Wang, Shuailong Liang, Yue Zhang, Xiaonan Li, and Tian Gao. 2019.
\newblock \href {https://doi.org/10.18653/v1/P19-1393} {Does it make sense? and why? a pilot study for sense making and explanation}.
\newblock In \emph{Proceedings of the 57th Annual Meeting of the Association for Computational Linguistics}, pages 4020--4026, Florence, Italy. Association for Computational Linguistics.

\bibitem[{Wei et~al.(2022{\natexlab{a}})Wei, Tay, Bommasani, Raffel, Zoph, Borgeaud, Yogatama, Bosma, Zhou, Metzler, Chi, Hashimoto, Vinyals, Liang, Dean, and Fedus}]{wei2022emergent}
Jason Wei, Yi~Tay, Rishi Bommasani, Colin Raffel, Barret Zoph, Sebastian Borgeaud, Dani Yogatama, Maarten Bosma, Denny Zhou, Donald Metzler, Ed~H. Chi, Tatsunori Hashimoto, Oriol Vinyals, Percy Liang, Jeff Dean, and William Fedus. 2022{\natexlab{a}}.
\newblock \href {https://openreview.net/forum?id=yzkSU5zdwD} {Emergent abilities of large language models}.
\newblock \emph{Trans. Mach. Learn. Res.}, 2022.

\bibitem[{Wei et~al.(2022{\natexlab{b}})Wei, Wang, Schuurmans, Bosma, Xia, Chi, Le, Zhou et~al.}]{wei2022chain}
Jason Wei, Xuezhi Wang, Dale Schuurmans, Maarten Bosma, Fei Xia, Ed~Chi, Quoc~V Le, Denny Zhou, et~al. 2022{\natexlab{b}}.
\newblock Chain-of-thought prompting elicits reasoning in large language models.
\newblock \emph{Advances in neural information processing systems}, 35:24824--24837.

\bibitem[{Yao et~al.(2024)Yao, Yu, Zhao, Shafran, Griffiths, Cao, and Narasimhan}]{yao2024tree}
Shunyu Yao, Dian Yu, Jeffrey Zhao, Izhak Shafran, Tom Griffiths, Yuan Cao, and Karthik Narasimhan. 2024.
\newblock Tree of thoughts: Deliberate problem solving with large language models.
\newblock \emph{Advances in Neural Information Processing Systems}, 36.

\bibitem[{Yuan et~al.(2023)Yuan, Yuan, Li, Dong, Tan, and Zhou}]{yuan2023scaling}
Zheng Yuan, Hongyi Yuan, Chengpeng Li, Guanting Dong, Chuanqi Tan, and Chang Zhou. 2023.
\newblock Scaling relationship on learning mathematical reasoning with large language models.
\newblock \emph{arXiv preprint arXiv:2308.01825}.

\bibitem[{Zelikman et~al.(2022)Zelikman, Wu, Mu, and Goodman}]{zelikman2022star}
Eric Zelikman, Yuhuai Wu, Jesse Mu, and Noah Goodman. 2022.
\newblock \href {https://openreview.net/forum?id=_3ELRdg2sgI} {{ST}ar: Bootstrapping reasoning with reasoning}.
\newblock In \emph{Advances in Neural Information Processing Systems}.

\bibitem[{Zhao et~al.(2023)Zhao, Joshi, Liu, Khalman, Saleh, and Liu}]{zhao2023slic}
Yao Zhao, Rishabh Joshi, Tianqi Liu, Misha Khalman, Mohammad Saleh, and Peter~J Liu. 2023.
\newblock Slic-hf: Sequence likelihood calibration with human feedback.
\newblock \emph{arXiv preprint arXiv:2305.10425}.

\bibitem[{Ziegler et~al.(2019)Ziegler, Stiennon, Wu, Brown, Radford, Amodei, Christiano, and Irving}]{ziegler2019finetuning}
Daniel~M. Ziegler, Nisan Stiennon, Jeffrey Wu, Tom~B. Brown, Alec Radford, Dario Amodei, Paul Christiano, and Geoffrey Irving. 2019.
\newblock Fine-tuning language models from human preferences.
\newblock \emph{arXiv preprint arXiv: 1909.08593}.

\end{thebibliography}
